\documentclass[10pt,journal,compsoc]{IEEEtran}%
\ifCLASSOPTIONcompsoc
  \usepackage[nocompress]{cite}
\else
  \usepackage{cite}
\fi

\ifCLASSINFOpdf
\else
\fi

\usepackage{graphicx}
\usepackage{amsmath,amssymb} 
\usepackage{color}
\usepackage[pagebackref=true,colorlinks,bookmarks=false,citecolor=blue,linkcolor=blue,urlcolor=blue]{hyperref}
\usepackage[english]{babel}
\usepackage{array}
\usepackage{algorithm} 
\usepackage{algorithmic} 
\usepackage{subfigure}
\usepackage{multirow}
\usepackage{slashbox}
\usepackage{url}
\usepackage{amsthm}

\usepackage{ragged2e}
\renewcommand{\raggedright}{\leftskip=0pt \rightskip=0pt plus 0cm}
\usepackage{makecell}
\usepackage{color}
\usepackage{booktabs}
\usepackage{enumitem}

\begin{document}
\title{Learning Disentangled  Representation for One-Shot Progressive Face Swapping}

\author{   
        Qi~Li,~\IEEEmembership{}
        Weining~Wang,~\IEEEmembership{}
        Chengzhong~Xu,~\IEEEmembership{}
        Zhenan~Sun,~\IEEEmembership{}
        and Ming-Hsuan Yang~\IEEEmembership{}%

\IEEEcompsocitemizethanks{

\IEEEcompsocthanksitem This work was supported in part by the National Key Research and Development Program of China under Grant No. 2022YFC3310400, the National Natural Science Foundation of China (Grant Nos. U23B2054, 62076240, 62102419, 62276263), the Beijing Municipal Natural Science Foundation (Grant No. 4222054), and the Macau FDCT (Grants Nos. 0123/2022/AFJ, 0081/2022/A2). (Corresponding authors: Weining Wang and Zhenan Sun.)

\IEEEcompsocthanksitem 
Qi Li and Zhenan Sun are with the New Laboratory of Pattern Recognition (NLPR), State Key Laboratory of Multimodal Artificial Intelligence Systems, Institute of Automation, Chinese Academy of Sciences, Beijing, 100190, China, and are also with the School of Artificial Intelligence, University of Chinese Academy of Sciences, Beijing 100049, China (email: qli@nlpr.ia.ac.cn; znsun@nlpr.ia.ac.cn).
\IEEEcompsocthanksitem 
Weining Wang is with the Laboratory of Cognition and Decision Intelligence for Complex Systems, Institute of Automation, Chinese Academy of Sciences, Beijing, 100190, China (email: weining.wang@nlpr.ia.ac.cn).
\IEEEcompsocthanksitem 
Chengzhong Xu is with the State Key Laboratory of Internet of Things for Smart City,  University of Macau, Macao, 999078, China (email: czxu@um.edu.mo).
\IEEEcompsocthanksitem 
Ming-Hsuan Yang is with the Department of Computer Science and
Engineering, University of California, Merced, CA, 95340, USA (email:mhyang@ucmerced.edu), and Department of Computer Science and Engineering, Yonsei University, Korea. 
}
}

\IEEEtitleabstractindextext{%
\raggedright{
\begin{abstract}
Although face swapping has attracted much attention in recent years, it remains a challenging problem. 
Existing methods leverage a large number of data samples to explore the intrinsic properties of face swapping without considering the semantic information of face images. 
Moreover, the representation of the identity information tends to be fixed, leading to suboptimal face swapping. In this paper, we present a simple yet efficient method named FaceSwapper, for one-shot face swapping based on Generative Adversarial Networks. 
Our method consists of a disentangled representation module and a semantic-guided fusion module. 
The disentangled representation module comprises an attribute encoder and an identity encoder, which aims to achieve the disentanglement of the identity and attribute information. 
The identity encoder is more flexible, and the attribute encoder contains more attribute details than its competitors. 
Benefiting from the disentangled representation, FaceSwapper can swap face images progressively. 
In addition, semantic information is introduced into the semantic-guided fusion module to control the swapped region and model the pose and expression more accurately. Experimental results show that our method achieves state-of-the-art results on benchmark datasets with fewer training samples. 
Our code is publicly available at \href{https://github.com/liqi-casia/FaceSwapper}{https://github.com/liqi-casia/FaceSwapper}.
\end{abstract}}

}

\maketitle

\IEEEdisplaynontitleabstractindextext

\IEEEpeerreviewmaketitle

\IEEEraisesectionheading{\section{Introduction}
\label{sec:introduction}}

\IEEEPARstart{F}{ace} swapping refers to transferring a face from a source face image to a target face image while preserving the pose, expression, lighting conditions, and other attributes of the target face image. 
With the rapid growth of methods for generating synthetic face images, face swapping has attracted much attention in recent years. 
As one of the most widely used techniques for face image synthesis, it facilitates a wide range of practical applications, such as appearance transfiguration in portraits, a potential data augmentation method for face analysis, and privacy protection.

Face swapping methods can be broadly categorized as 3D model-based and Generative Adversarial Networks (GANs)-based methods.
3D model-based methods usually transfer face images using a 3D morphable model (3DMM)~\cite{blanz1999morphable,tran2018nonlinear}, where a dense point-to-point correspondence is established between the source face and the target face.
A pipeline for 3D model-based face swapping methods can be formulated as follows.
The first procedure is to fit a 3DMM to face images to obtain the approximate shape, texture, and lighting conditions.
The shape can be further represented as a combination of identity and expression coefficients. 
The visible parts of the source and target faces are then segmented from their context and occlusions. 
The source and target faces are swapped according to the estimated 3D shape coordinates.
Finally, the replaced face texture is re-rendered and blended with the target face image.
The advantage of 3D model-based methods is that the pose, expression, and lighting conditions are disentangled well since the texture branch focuses on appearance changes, such as wrinkles, shadows, and shading, while geometric details are displayed mainly in the shape branch.
Thus, 3D model-based methods can handle face images with large pose variations, exaggerated expressions, and drastic lighting conditions. 
Another advantage of 3D model-based methods is that they can be easily extended to process high-resolution face images.
However, there are also some drawbacks. 
For example, it is difficult to accurately estimate the geometries and textures of face images. 
In addition, the optimization for these methods is susceptible to local minimal values, and the computational cost is usually higher than other methods.

Numerous GAN models have been proposed for generating synthetic image samples~\cite{goodfellow2014generative,karras2019style,karras2020analyzing,karras2021alias}.
These models have also been widely explored for face swapping and perform well against existing methods by generating realistic face images~\cite{nirkin2019fsgan,nitzan2020face,chen2020simswap,zhu2021one}. 
Several open-source projects based on GANs for face swapping have recently been proposed. DeepFake~\cite{DeepFake-url} utilizes GANs to make face swapping widely accessible to the public and attracted much attention.
DeepFaceLab~\cite{petrov2020deepfacelab} is another open-source project providing an easy-to-use pipeline for face swapping.
In addition, some commercial mobile applications for face swapping, such as ZAO and FaceApp, have been developed.

\begin{figure*}[hptb]
\begin{center}
\includegraphics[width=1.0\linewidth]{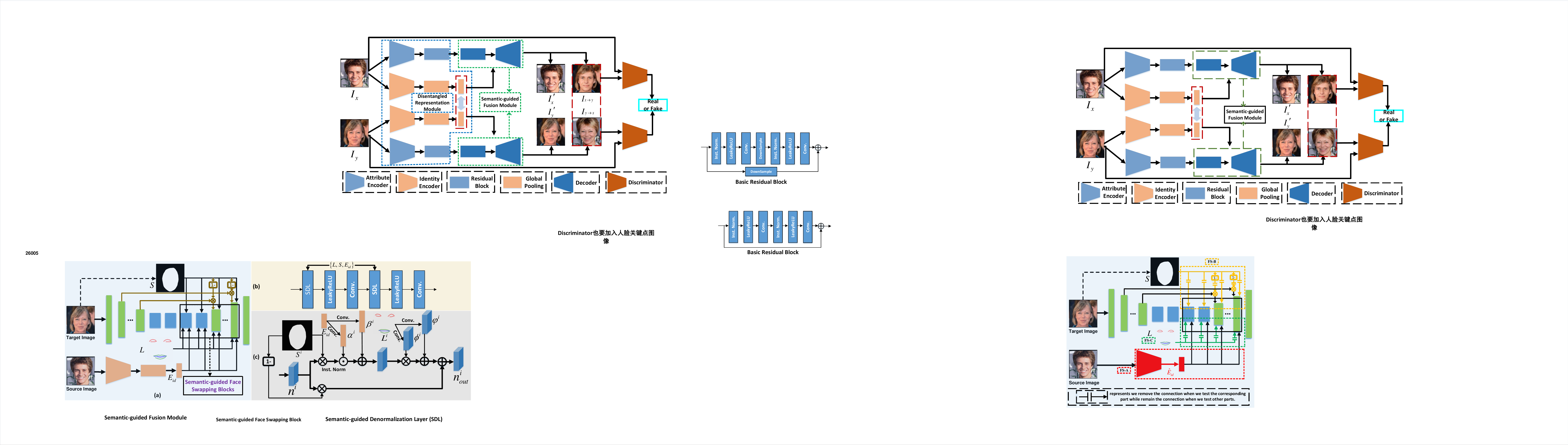}
   \caption{\textbf{Main components of FaceSwapper.} It consists of two generators and two discriminators. The generator consists of a disentangled representation module (DRM) and a semantic-guided fusion module (SFM). DRM (blue dashed lines) comprises an attribute encoder and an identity encoder. SFM (green dashed lines) takes the output of the attribute encoder and the output of the identity encoder as inputs and synthesizes an output image that combines the attributes of the former and the identity of the latter. The discriminator is a conditional discriminator trained to distinguish synthetic images from real images while preserving landmark information. }
\label{fig:framework}
\end{center}
\end{figure*}

While significant advances have been made for face swapping, numerous challenges remain.
First, to enable realistic face swapping, existing methods usually require many images from the same identity for appearance modeling. 
One drawback of these methods is that they have poor generalization ability and are difficult to adapt to arbitrary faces. 
Second, it is difficult to completely disentangle the identity information from the attribute information, which leads to less controllability of the swapped faces.
Furthermore, semantic information is exploited in a limited way in previous methods. 
As a result, a large number of images must be leveraged to train an attention-based model or a refinement model for face swapping. 
Considering the above context, we focus on a challenging task, one-shot face swapping, in which we are given only one face image each from the source and target identities in both training and testing.
We present a simple yet efficient method, FaceSwapper, for one-shot face swapping.
FaceSwapper generalizes and adapts to arbitrary faces well.
Figure~\ref{fig:framework} shows an overview of FaceSwapper.

The main contributions of this work are:
\begin{itemize}
\item We present an end-to-end framework for one-shot progressive face swapping based on GANs, which 
consists of a disentangled representation module and a semantic-guided fusion module.
FaceSwapper can swap faces progressively with more controllable and diverse results. 

\item The disentangled representation module, which contains an identity encoder and an attribute encoder, is exploited to disentangle the identity and attribute information automatically. 
Different from previous works,  the parameters of the identity encoder are learned fully automatically during the training process.
As such, the identity information can be injected into FaceSwapper with more controllability. 
In addition, the attribute encoder in our method preserves more details of the attributes than its competitors.

\item Semantic information is introduced into FaceSwapper to accurately synthesize the results and avoid influences from complex backgrounds. 
This information is considered a modulating coefficient injected into the normalization layers of the semantic-guided fusion module. It can be seamlessly fitted to the synthesized results without inconsistency, thereby allowing the generation of high-fidelity swapped images. 

\item Extensive experimental results show that FaceSwapper achieves state-of-the-art results on benchmark datasets with only 30,000 training samples. 
In addition, we present a cross-dataset experiment in the appendix to show the superiority of our method in terms of its generalization ability.
\end{itemize}

\section{Related Work}
\label{sec:relatedwork}

We discuss prior work based on 3D and GAN models for face swapping as well as reenactment in this section.

\textit{3D face representation} methods are developed based on explicit parametric representations of face geometry.
Blanz~\emph{et al.}~\cite{blanz2004exchanging} propose an algorithm to estimate the 3D shape, texture and rendering parameters for both source and target images. 
The swapped face image can then be obtained by rendering the face that was reconstructed from the source face image with the rendering parameters from the target face image.
One drawback of this method is that it requires manual interaction, which is tedious and time-consuming.
Bitouk~\emph{et al.}\cite{bitouk2008face} present a three-stage method for swapping faces with a large-scale face image database. 
The pose and expression of the source and target faces must be similar.
Lin~\emph{et al.}\cite{lin2012face,lin2014pose} introduce a method for building a personalized 3D
head model from a frontal face image, which can be rendered in any pose to match the target face. 
The source face and target face are then swapped based on the proposed personalized 3D head models. 
Color transfer and multiresolution techniques are also applied to make the results seamless and natural.
Cheng~\emph{et al.}~\cite{cheng20093d} utilize a 3DMM for face swapping in video.
One assumption of this method is that the acquisition  process of the source subject is under control. 
Dale~\emph{et al.}~\cite{dale2011video} present a method for replacing faces in videos. 
A 3D multilinear model is used to track the face in both the source and the target videos. Then, the source video is warped to the target video in space and time. 
Finally, the videos are seamlessly blended by computing a spatial-temporal seam.
In \cite{nirkin2018face}, 2D facial landmarks are first localized to compute the 3D pose, shape and expression of a face using a deep model~\cite{tran2018nonlinear}.
Faces are then segmented from their backgrounds and occlusions using a trained fully convolutional neural network. 
Finally, faces are swapped using aligned 3D shapes as proxies.
One limitation of this method is that it is not end-to-end trainable.

\textit{GANs-based} methods have been shown to synthesize more realistic results than conventional face swapping schemes.
Bao~\emph{et al.} propose CVAE-GAN~\cite{bao2017cvae}, which combines a variational autoencoder with a generative adversarial network for face swapping. 
One limitation of this method is that it needs hundreds of images for each person.
Korshunova~\emph{et al.}~\cite{korshunova2017fast} formulate face swapping as a neural style transfer problem.
However, this method requires numerous images for a single person.
Natsume~\emph{et al.}~\cite{natsume2018rsgan} design a neural network to encode the face appearance and hair appearance into separate latent space representations with two variational autoencoders.
The generative adversarial network is then used to generate a natural face image from the latent space representations.
A high-resolution neural face swapping method is proposed in~\cite{naruniec2020high}.
First,  a neural network is employed to encode high-resolution images. 
Then, a light- and contrast-preserving network is adopted to ensure the consistency of the target and swapped images.
One common drawback of the above face swapping methods is that dozens of images or more are required to obtain a realistic synthesized result.

Nirkin~\emph{et al.}~\cite{nirkin2019fsgan} develop a novel recurrent neural network (RNN)-based approach for subject agnostic face swapping named a face swapping GAN (FSGAN) that can be applied to faces of different subjects without requiring subject-specific training. 
This scheme consists of a recurrent reenactment generator, a segmentation generator, an inpainting generator, and a blending generator. 
It breaks large pose changes into small steps to handle large pose variations and then interpolates between the closest available images.
An identity-preserving GAN (IPGAN) is developed in~\cite{bao2018towards} to disentangle the identity and attributes of faces and then synthesize faces from the recombined identity and attributes.
Li~\emph{et al.}~\cite{li2020advancing} propose FaceShifter, a novel two-stage algorithm, to obtain high-fidelity face swapping results.
In~\cite{zhu2021one}, Zhu~\emph{et al.} present the first megapixel level method for one-shot face swapping. It consists of three stages for high-fidelity face swapping. Firstly, HieRFE is employed to encode faces into hierarchical representations. Then FTM is carefully designed to transfer the identity from the source image to the target image. Finally, StyleGAN2~\cite{karras2020analyzing} is used to synthesize the swapped face image.
Xu~\emph{et al.}~\cite{xu2022high} also propose a high-resolution face swapping method based on the inherent prior knowledge of StyleGAN. 
The attributes are categorized into structure and appearance ones, and then they are transferred separately in the disentangled latent space.
To better preserve the identity consistency of the generated face, Xu~\emph{et al.}~\cite{xu2022region} 
explicitly model the local facial features besides the global representations to perform feature interaction more finely.
One limitation of the above methods is that many images are needed to train the algorithms, and are not end-to-end trainable.

\textit{Face reenactment} methods aim to transfer the pose and expression of the source image to the target image while the identity remains intact. 
A user-assisted technique is proposed in~\cite{pighin2006synthesizing}~to recover the camera poses and facial geometry, and then the 3D shape is used to transfer different expressions.
Thies~\emph{et al.}~\cite{thies2015real} present an end-to-end approach for real-time face reenactment.
One limitation of this method is that it relies on depth data.
In subsequent work, Thies~\emph{et al.}~\cite{thies2016face2face} propose a real-time
face reenactment method based on RGB videos. 
The facial expressions of both the source and target videos are tracked using a dense photometric consistency measure. 
Then, reenactment is achieved by fast and efficient deformation transfer.
GANs have also been successfully applied for face reenactment.
A novel learning-based framework for face reenactment, ReenactGAN, is proposed in~\cite{wu2018reenactgan}. 
Due to the usage of a target-specific decoder, this method cannot be used for unknown identities.
On the other hand, few-shot face reenactment methods have been
proposed to reenact unknown identities.
In~\cite{zakharov2019few}, a few-shot adversarial learning approach for face reenactment is presented. 
The advantage of this method is that it can initialize the parameters of deep neural networks in a person-specific way due to the usage of meta-learning.
A one-shot face reenactment method is proposed by Zhang~\emph{et al.}~\cite{zhang2019one}.
The target face appearance and the source shape are first projected into latent spaces with their corresponding encoders. 
Then a shared decoder is used to inject the latent representations to obtain the final reenactment results.
Ha~\emph{et al.}~\cite{ha2019marionette} present a method to reenact the faces of unseen targets in a few-shot manner while preserving identity without fine-tuning. 
Burkov~\emph{et al.}~\cite{burkov2020neural} propose a simple reenactment method driven by a latent pose representation, and 
Nitzan~\emph{et al.}~\cite{nitzan2020face} present a scheme to disentangle face identity from all other facial attributes without data-specific supervision.

\section{Method}
\label{sec:overall-framework}
Assume we have two input images, represented as ${I_x}$ and ${I_y}$. 
Face swapping aims to synthesize a new face image ${I_{x \to y}}$ that has the same identity as ${I_x}$ (the source image) and the same attributes, e.g., pose, expression, lighting, and background, as ${I_y}$ (the target image). 
Face swapping is highly underconstrained due to the shortage of ground-truth data samples.
In this paper, we exploit another symmetric network to alleviate this problem.
That is, we synthesize a face image ${I_{y \to x}}$ that has the same identity as ${I_y}$ (the source image) and the same attributes as ${I_x}$ (the target image).
The proposed face swapping method is based on the symmetric network structure.

Specifically, an adversarial learning framework is constructed in our method, which
consists of a generator $G$ and a discriminator $D$.
The generator $G$ consists of a Disentangled Representation Module (DRM), named $E{}_{enc}$,  and a Semantic-guided Fusion Module (SFM), denoted $E{}_{dec}$.
$E{}_{enc}$ consists of two networks: an attribute encoder network ${E_{attr}}$ and an identity encoder network ${E_{id}}$. $E{}_{dec}$ contains only one network: a semantic-guided fusion decoder network ${G_{f}}$.
Given an input image ${I_x}$ or ${I_y}$, ${E_{attr}}$ is trained to encode the attribute information (e.g., pose, expression, lighting, and background).
In contrast to ${E_{attr}}$, ${E_{id}}$ takes ${I_x}$ or ${I_y}$ as input and is explored to extract the identity information. 
Finally, ${G_{f}}$ combines the identity information of ${I_x}$ (or ${I_y}$) with the attribute information of ${I_y}$ (or ${I_x}$) and synthesizes the swapped image ${I_{x \to y}}$ (or ${I_{y \to x}}$).
Here,  $D$ is a discriminator trained to distinguish synthetic and real images. 
Figure~\ref{fig:framework} shows the overall framework of our method.

\subsection{Disentangled Representation Module}

For face swapping, separating the attribute information from the identity information in a given face image is an important problem. 
Motivated by recent advances in style transfer~\cite{gatys2016image,huang2018multimodal}, we exploit DRM, which contains a ``content'' encoder and a ``style'' encoder, to learn disentangled representations for face swapping. 
We formulate identity information as ``style'' information, while attribute information can be seen as ``content'' information.
Both types can be optimized during the training process.
In this work, DRM disentangles the representation of a face image with an attribute encoder ${E_{attr}}$ and an identity encoder ${E_{id}}$.
In addition, we use residual blocks~\cite{he2016deep,he2017mask}
as the basic network architecture units.
Note that we choose instance normalization rather than batch normalization in this paper because instance normalization normalizes each sample to a single ``style''. In contrast,  batch normalization normalizes a batch of samples to be centered around a single ``style''. 
Face swapping needs to preserve the identity or attributes of each sample, which means instance normalization is better for generating images with vastly different identities.

We next present the detailed network structures of ${E_{attr}}$ and ${E_{id}}$.
Thanks to recent advances in face recognition~\cite{schroff2015facenet,deng2019arcface}, the identity information can be partially represented by a vector.
In  comparison, the attribute information contains much identity-irrelevant information, which is difficult to represent using a vector.
Thus, residual bottleneck blocks, which consist of  multilevel feature maps, are appended at the end of the attribute encoder to represent the attribute information.
Specifically, ${E_{attr}}$ contains five downsampling and two bottleneck residual blocks.
Compared with the downsampling residual block, the bottleneck residual block has no downsampling operations. 
The identity encoder ${E_{id}}$ has six downsampling residual blocks, and then a convolutional layer and a fully connected layer are appended at the end.

Given the input images $I_x$ and $I_y$, ${E_{id}}$ is trained to learn the latent identity codes ${E_{id}}\left( {{I_x}} \right)$ and ${E_{id}}\left( {{I_y}} \right)$.
Specifically, ${E_{attr}}$ is trained to learn the latent attribute codes ${E_{attr}}\left( {{I_x}} \right)$ and ${E_{attr}}\left( {{I_y}} \right)$.
To transfer the identity information of the face image $I_x$ to the face image $I_y$, we recombine its identity code ${E_{id}}\left( {{I_x}} \right)$ with the attribute code ${E_{attr}}\left( {{I_y}} \right)$, and then send the combined code to SFM to produce the final swapped face image ${I_{x \to y}}$.
Since face swapping is a highly underconstrained problem, it is difficult to disentangle such information accurately without other constraints.
Naturally, we can have another symmetric network to regulate DRM.
That is, to transfer the identity information of the face image $I_y$ to the face image $I_x$, we recombine its identity code ${E_{id}}\left( {{I_y}} \right)$ with the attribute code ${E_{attr}}\left( {{I_x}} \right)$ and then send the combined code to SFM  to produce the final swapped face image ${I_{y \to x}}$.
To capture more details of the target image, the skip connection~\cite{ronneberger2015u} is exploited in the attribute encoder, which directly concatenates the feature maps at different resolutions in the encoder and the corresponding ones in the decoder.

\begin{figure*}[tp]
\begin{center}
\includegraphics[width=.8\linewidth]{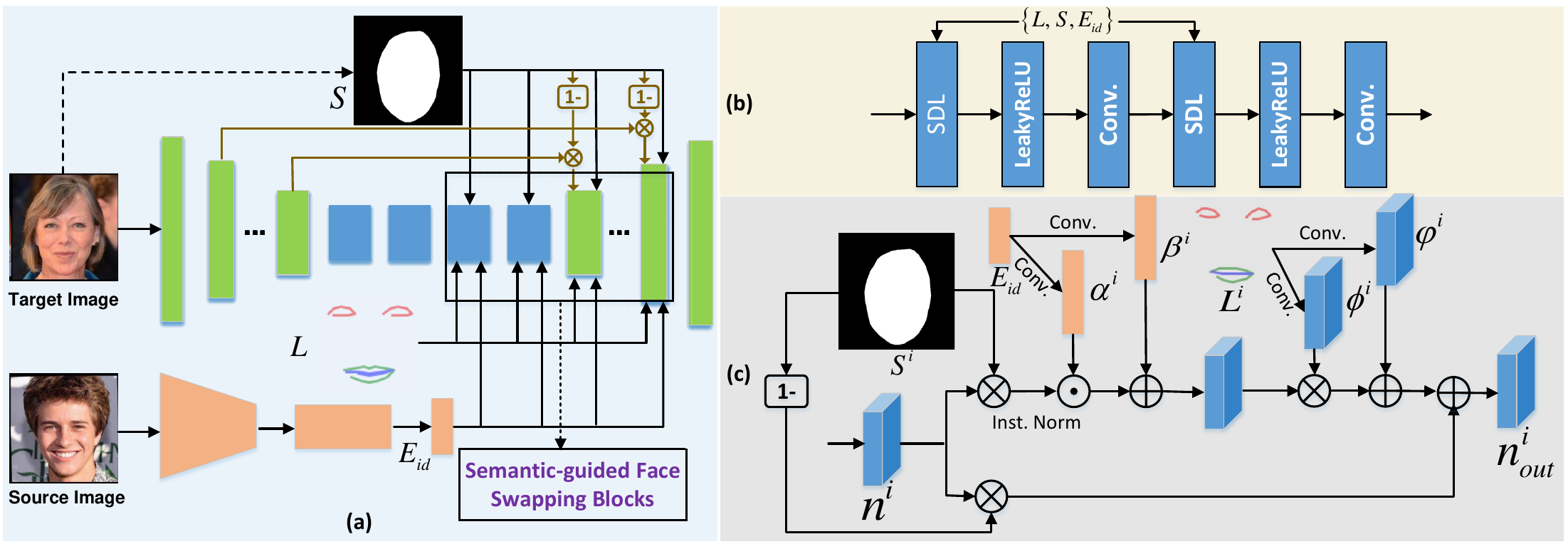}
\vspace{-4mm}
\caption{\textbf{Overview of the semantic-guided fusion module.} (a) The semantic-guided fusion module integrates the identity and attribute information using multiple semantic-guided face swapping blocks. (b) The specific network structure of the semantic-guided face swapping block, which is built on the semantic-guided denormalization layer. (c) Details of the semantic-guided denormalization layer.}
\label{fig:framework-SFM}
\end{center}
\end{figure*}

\subsection{Semantic-guided Fusion Module}
SFM takes an identity code and an attribute code as inputs and synthesizes an output image that recombines the identity of the former and the attributes of the latter.
Since the processes to produce the synthesized images ${I_{x \to y}}$ and ${I_{y \to x}}$ are similar, in the following, we take ${I_{x \to y}}$ as an example to illustrate this module for brevity.
In contrast to prior work, a semantic-guided fusion module is exploited to inject the identity code into the generator. 
Our method assumes that the swapped face image should be the same as the target image except for the face area. 
Thus, a semantic mask, generated by a face parsing algorithm~\cite{yu2018bisenet}, is adopted to combine the identity and attribute codes.
Figure~\ref{fig:framework-SFM}(a) presents an overview of SFM.
SFM comprises two bottleneck and five upsampling semantic-guided face swapping blocks.
Compared with the bottleneck block, the upsampling block has another upsampling operation in the middle of the block. 
Figure~\ref{fig:framework-SFM}(b) shows the proposed semantic-guided face swapping block, which is similar to the basic residual block
except that the normalization layer is replaced by a semantic-guided denormalization layer (SDL).
Similar to DRM, instance normalization is utilized in SDL, which prevents instance-specific mean and covariance shifts and simplifies the learning process.

Figure~\ref{fig:framework-SFM}(c) shows the details of SDL.
Given an input batch of $B$ samples, the activation of the $i$-th layer of SDL can be
represented as ${n^i} \in {\mathbb{R}^{B \times {C^i} \times {H^i} \times {W^i}}}$,
which is a four-dimensional tensor, where $C^i$ denotes the number of channels in the $i$-th layer, and $H^i$ and $W^i$ represent the height and width of the activation map in the $i$-th layer, respectively.
Let $S \in {\left\{ {0,1} \right\}^{B \times 1 \times H \times W}}$ denote the original binary mask generated by a face parsing algorithm.
Since each face swapping block operates at a different scale, we downsample $S$ to match the spatial resolution of the $i$-th layer.
We resize $S$ to the same spatial resolution as ${n^i}$ and repeat ${C^i}$ times to obtain the binary mask of the $i$-th layer
${S^i} \in {\left\{ {0,1} \right\}^{B \times {C^i} \times {H^i} \times {W^i}}}$. Each entry in ${S^i}$ indicates whether it relates to the swapped face area.
Let ${n_{bcxy}^i}$ denote the $bcxy$-th element of ${n^i}$ and $S_{bcxy}^i$ represent the $bcxy$-th element of ${S^i}$, where $b \in B$ is the index of the image in the batch, $c \in {C^i}$ is the index of the feature channel, and $x \in {H^i}$ and $y \in {W^i}$ span the spatial dimensions.

In the proposed method, ${n^i}$ is normalized to zero mean and unit deviation for the swapped face area by instance normalization:
\begin{equation}
\bar n_{bcxy}^i{\rm{ = }}\frac{{n_{bcxy}^i \otimes S_{bcxy}^i - \mu _{^{bc}}^i}}{{\sigma _{bc}^i}},
\end{equation}
where $\otimes$ denotes elementwise multiplication, $\bar n_{bcxy}^i$ is the $bcxy$-th element of $\bar n^i$  after normalization. 
Here, ${\mu _{^{bc}}^i}$ and ${\sigma _{bc}^i}$ are computed across spatial dimensions
independently for each channel and each sample. 
To integrate the identity code ${E_{id}}\left( {{I_x}} \right)$ into the attribute code ${E_{attr}}\left( {{I_y}} \right)$, we compute the output activation by denormalizing  $\bar n^i$ as follows:
\begin{equation}
\label{eq:id-injection}
{m^i} = {\alpha ^i}\left( {{E_{id}\left( {{I_x}} \right)}} \right) \odot {\bar n^i} + {\beta ^i}\left( {{E_{id}\left( {{I_x}} \right)}} \right),
\end{equation}
where $\odot$ represents channel-wise multiplication, ${m^i} \in {\mathbb{R}^{B \times {C^i} \times {H^i} \times {W^i}}}$ is the activation after the denormalization process, and ${\alpha ^i}\left( {{E_{id}\left( {{I_x}} \right)}} \right)$ and ${\beta ^i}\left( {{E_{id}\left( {{I_x}} \right)}} \right)$
are two learned modulation parameters of SDL convolved from $E_{id}$, which can be implemented using a simple two-layer fully connected network. 
Note that both ${\alpha ^i}\left( {{E_{id}\left( {{I_x}} \right)}} \right)$ and ${\beta ^i}\left( {{E_{id}\left( {{I_x}} \right)}} \right)$ are one-dimensional  vectors that share the same channel dimension as $\bar n^i$.
Eqn.~\ref{eq:id-injection} means that we simply scale $\bar n^i$ with ${\alpha ^i}\left( {{E_{id}\left( {{I_x}} \right)}} \right)$ and then shift it by ${\beta ^i}\left( {{E_{id}\left( {{I_x}} \right)}} \right)$.

With Eqn.~\ref{eq:id-injection}, we can inject the identity information of
$I_x$ into $I_y$. 
However, since the spatial information of the face area provided by ${E_{attr}}\left( {{I_y}} \right)$ is limited, our method only learns to encode the appearance information of $I_x$ while omitting the pose and expression information from $I_y$ (e.g., the states of the mouth and eye).
Thus, learning precise pose and expression information from $I_y$ is a challenging problem.
Facial landmarks are widely exploited in face expression synthesis~\cite{song2018geometry} and face reenactment~\cite{zakharov2019few,zhang2019one}. 
Similar to these methods, we use facial landmarks to transmit spatial information into the decoder to guide learning. 
Facial landmarks enable the pose and expression to be learned visually and can be adjusted with different intensities. 
Their usage provides an intuitive and efficient way to specify the pose and expression from the target image.

The landmark features convolved from the landmark images are used to modulate the input features. 
Specifically, we use an off-the-shelf face alignment algorithm~\cite{bulat2017far} to locate the landmarks of a given image. 
The landmark features are then rasterized into a three-channel image using distinct colors to connect landmarks of different facial components (e.g., eyes and mouth) with line segments.
We denote the landmark images resulting from the target images as ${L} \in {\mathbb{R}^{B \times 3 \times H \times W}}$. ${L}$ can be resized to the same spatial resolution with ${m^i}$, represented as ${L^i} \in {\mathbb{R}^{B \times 3 \times {H^i} \times {W^i}}}$.
Unlike Eqn.~\ref{eq:id-injection}, we propose using the spatially-adaptive denormalization method~\cite{park2019semantic} to inject ${L^i}$ into ${m^i}$:
\begin{equation}
{p^i} = {\phi ^i}\left( {{L^i}} \right) \otimes {m^i} + {\varphi ^i}\left( {{L^i}} \right),
\end{equation}
where $p^i$ is the activation after the denormalization procedure, and
$\phi ^i\left( {{L^i}} \right)$ and $\varphi ^i\left( {{L^i}} \right)$ are the learned scale and bias modulation parameters which share the same tensor dimension as $m^i$. 
They are convolved from the input ${L^i}$ and vary with respect to spatial locations. The convolution consists of a simple two-layer convolutional neural network.
Finally, a simple fusion operation is used to combine the inner face area and the background. 
The final output can be represented as follows:
\begin{equation}
n_{out}^i = {n^i} \otimes \left( {1 - {S^i}} \right) + {p^i}.
\end{equation}

\subsection{Loss Function}
Our loss function comprises an identity preservation loss to keep the personalized features of the source image, an attribute preservation loss to maintain the attributes of the target image, a reconstruction loss that ensures that the encoders and decoders are inverses of each other, and an adversarial loss that encourages the distribution of the synthetic images to be indistinguishable from that of the real images.

\subsubsection{Identity and Attribute Preservation Loss}
Based on the proposed model, we encourage the swapped image ${I_{x \to y}}$ to have the same identity as ${I_x}$ and the same attributes as ${I_y}$. 
In this paper, the identity preservation loss is employed to minimize the distance between ${I_{x \to y}}$ and ${I_x}$ in the feature space, which embeds personalized identity-related information.
Specifically, ${I_{x \to y}}$ and ${I_x}$ are first resized to a resolution of 112 x 112, and then these images are sent to ${\psi _{id}}$, which is a pretrained state-of-the-art face recognition model~\cite{deng2019arcface}. 
We use ${\psi _{id}}$ as the feature extractor, and its parameters are fixed during optimization.
The identity preservation loss is defined based on the activations of the last fully connected layers of ${\psi _{id}}$:
\begin{equation}
\label{eq:id-loss}
{\mathcal{L}_{id}^{x \to y}} = 1 - \frac{{{\psi _{id}}{{\left( {{I_x}} \right)}^T}{\psi _{id}}\left( {{I_{x \to y}}} \right)}}{{{{\left\| {{\psi _{id}}\left( {{I_x}} \right)} \right\|}_2}{{\left\| {{\psi _{id}}\left( {{I_{x \to y}}} \right)} \right\|}_2}}},
\end{equation}
where ${\left\|  \bullet  \right\|_2}$ denotes the ${\ell _2}$-norm of a vector.
Eqn.~\ref{eq:id-loss} is employed to measure the cosine distance of ${I_{x \to y}}$ and ${I_x}$ in the feature space.

For the attribute preservation loss, one common practice is using pretrained attribute classifiers to minimize the distance between the input images and the synthetic images in the feature space.
However, defining an exact attribute classifier for face swapping is difficult (e.g., pose, expression and  lighting). 
In this paper, we exploit multilevel feature maps of SFM to represent the attributes with a component named the attribute extractor ${\psi _{attr}}$. 
We require that the swapped face image ${I_{x \to y}}$ and the target image ${I_y}$ have the same attribute feature maps outside the face area.
Suppose that $T$ is the total number of feature maps in SFM; then, the attribute preservation loss can be calculated as the pixel level ${\ell _2}$-distance between ${I_{x \to y}}$ and ${I_y}$:
\begin{equation}
\mathcal{L}_{attr}^{x \to y} = \sum\limits_{t = 1}^T {\left\| {\left( {\psi _{attr}^t\left( {{I_{x \to y}}} \right) - \psi _{attr}^t\left( {{I_y}} \right)} \right) \otimes \left( {1 - {S^t}} \right)} \right\|_2^2},
\end{equation}
where $\psi _{attr}^t\left(  \bullet  \right)$ denotes the $t$-th level feature maps, and ${S^t}$ represents the resized binary mask, which has the same spatial resolution as $\psi _{attr}^t\left(  \bullet  \right)$.
Similarly, we have the following loss function for ${I_{y \to x}}$:
\begin{equation}
\label{eq:id-loss-symmetry}
{\mathcal{L}_{id}^{y \to x}} = 1 - \frac{{{\psi _{id}}{{\left( {{I_y}} \right)}^T}{\psi _{id}}\left( {{I_{y \to x}}} \right)}}{{{{\left\| {{\psi _{id}}\left( {{I_y}} \right)} \right\|}_2}{{\left\| {{\psi _{id}}\left( {{I_{y \to x}}} \right)} \right\|}_2}}},
\end{equation}
\begin{equation}
\mathcal{L}_{attr}^{y \to x} = \sum\limits_{t = 1}^T {\left\| {\left( {\psi _{attr}^t\left( {{I_{y \to x}}} \right) - \psi _{attr}^t\left( {{I_x}} \right)} \right) \otimes \left( {1 - {S^t}} \right)} \right\|_2^2}.
\end{equation}
The identity and attribute preservation loss can be summarized as:
\begin{equation}
{\mathcal{L}_{id}} = \mathcal{L}_{id}^{x \to y} + \mathcal{L}_{id}^{y \to x},
\end{equation}
\begin{equation}
{\mathcal{L}_{attr}} = \mathcal{L}_{attr}^{x \to y} + \mathcal{L}_{attr}^{y \to x}.
\end{equation}

\subsubsection{Reconstruction Loss}  

One challenging problem for face swapping is that no ground-truth supervision is provided for training.
To encourage the network to generate realistic and meaningful face images, a constraint on the reconstruction of the original images should be accounted for in the network.
Therefore, a reconstruction loss is
employed to penalize the difference between the input image and its reconstruction:
\begin{equation}
\begin{array}{l}
\mathcal{L}_{rec}^x = \left\| {E{}_{dec}\left( {{E_{id}}\left( {{I_x}} \right),{E_{attr}}\left( {{I_x}} \right)} \right) - {I_x}} \right\|_2^2,\\
\mathcal{L}_{rec}^y = \left\| {E{}_{dec}\left( {{E_{id}}\left( {{I_y}} \right),{E_{attr}}\left( {{I_y}} \right)} \right) - {I_y}} \right\|_2^2.
\end{array}
\end{equation}
The overall reconstruction loss can be calculated as:
\begin{equation}
\begin{array}{l}
{\mathcal{L}_{rec}} = \mathcal{L}_{rec}^x + \mathcal{L}_{rec}^y.
\end{array}
\end{equation}

\subsubsection{Adversarial Loss}
To make the generated images more photorealistic, an adversarial loss is employed to encourage the images to be indistinguishable from real face images.
More specifically, our discriminator network $D$ takes pairwise samples as the input,
i.e., swapped face images with the landmark images of target faces or real
face images with their corresponding landmark images.
As a result, the discriminator can not only guide the generator
to produce realistic face images but also promote face images
containing the same landmarks as the target images.

Therefore, data
pairs of real faces with their corresponding landmarks are considered positive samples to train the discriminator network.
Data pairs of synthetic faces with the corresponding landmarks are considered negative samples.
Formally, the objective function can be denoted as:
\begin{align}
 \mathcal{L}_{adv}^{x \to y} = &\:{\mathbb{E}_{\left( {{I_x},{I_y},{L_y}} \right) \sim P\left( {{I_x},{I_y},{L_y}} \right)}}\left[ {\log \left( {1 - D\left( {G^{x \to y},{L_y}} \right)} \right)} \right] \nonumber\\
 + &\:{\mathbb{E}_{\left( {I_y,{L_y}} \right) \sim P\left( {I_y,{L_y}} \right)}}\left[ {\log D\left( {I_y,{L_y}} \right)} \right],
 \label{eq:adv-loss-x-y}
\end{align}
\begin{align}
 \mathcal{L}_{adv}^{y \to x} = &\:{\mathbb{E}_{\left( {{I_x},{I_y},{L_x}} \right) \sim P\left( {{I_x},{I_y},{L_x}} \right)}}\left[ {\log \left( {1 - D\left( {G^{y \to x},{L_x}} \right)} \right)} \right] \nonumber\\
 + &\:{\mathbb{E}_{\left( {I_x,{L_x}} \right) \sim P\left( {I_x,{L_x}} \right)}}\left[ {\log D\left( {I_x,{L_x}} \right)} \right],
 \label{eq:adv-loss-y-x}
\end{align}
where
\begin{equation}
\begin{array}{l}
{G^{x \to y}}{\rm{ = }}{E_{dec}}\left( {{E_{id}}\left( {{I_x}} \right),{E_{attr}}\left( {{I_y}} \right)} \right),\\
{G^{y \to x}}{\rm{ = }}{E_{dec}}\left( {{E_{id}}\left( {{I_y}} \right),{E_{attr}}\left( {{I_x}} \right)} \right).
\end{array}
\end{equation}
The  adversarial loss can be summarized as:
\begin{equation}
\mathcal{L}_{adv} = \mathcal{L}_{adv}^{x \to y} + \mathcal{L}_{adv}^{y \to x}.
\end{equation}

\subsubsection{Total Loss}
The final objective function is a combination of the above loss functions:
\begin{equation}
{\mathcal{L}_{total}} = {\lambda _{id}}{\mathcal{L}_{id}} + {\lambda _{attr}}{\mathcal{L}_{attr}} + {\lambda _{rec}}{\mathcal{L}_{rec}} + {\mathcal{L}_{adv}}.
\end{equation}
where $\lambda _{id}$, $\lambda _{attr}$ and $\lambda _{rec}$ are the weights that control the relative importance of each term.
Finally, the generator $G$ and the discriminator $D$ can be jointly trained by optimizing the following formula:
\begin{equation}
\mathop {\min }\limits_G \mathop {\max }\limits_D {\mathcal{L}_{total}}.
\end{equation}

\subsection{Relation to Previous Work}
\label{sec:relation-previous-works}
In this section, we discuss the differences between our model and
several state-of-the-art face synthesis methods: FSLSD~\cite{xu2022high}, RAFSwap~\cite{xu2022region}, MegaFace~\cite{zhu2021one}, 
FaceShifter~\cite{li2020advancing}, LSM~\cite{nitzan2020face}, IPGAN~\cite{bao2018towards}.
First, the proposed model is different from these methods in the disentanglement of the identity and attributes information.
Although the identity information is used in previous methods~\cite{li2020advancing,bao2018towards,nitzan2020face}, all of them employ a pre-trained face recognizer to extract the identity information, which lacks controllability (the identity information is fixed and frozen during training).
Recent high-resolution face swapping methods~\cite{xu2022high,xu2022region,zhu2021one} leverage the inherent prior knowledge of StyleGAN to extract identity information. However, it is difficult to disentangle the identity and attribute information completely due to the complex representation of StyleGAN. Thus how to accurately control the swapped face images accurately is a challenging problem for these methods.
Benefiting from the symmetrical design of the proposed framework, both the identity and attribute information are learned automatically in our method.
Furthermore, the semantic information is exploited in our method to aesthetically
and meticulously synthesize the face images to present the swapping effects.
Although previous methods have also exploited semantic information for high-resolution face swapping~\cite{xu2022high,xu2022region}, there is a big difference between these methods and our method regarding how to utilize the semantic information. 
The detailed architecture of our generator and the combination of the identity and attribute information are also different from previous methods.
In addition, unlike most other methods using a large amount of training data samples, our method can be trained with a limited number of training samples (about 30,000 images).
Taking FaceShifter as an example,  it is trained with a total of 2.7M images, approximately 90 times more than used to train our method. 

\section{Experiments}
\label{sec:experiments}

\subsection{Datasets and Experimental Settings}

In this subsection, we first describe the evaluation datasets and then present the implementation details of our algorithm.

\begin{itemize}
\item \textbf{CelebFaces Attributes Dataset (CelebA)}:
CelebA is a large-scale face dataset with more than 200,000 celebrity images~\cite{liu2015faceattributes}. It contains face images with large pose variations and background clutter. 
It has great diversity, a large number of samples and rich annotations, including 10,177 identities, 5 landmark locations and 40 binary attributes.

\item \textbf{CelebA-HQ}:
CelebA-HQ is originally selected from the CelebA dataset and further post-processed by the method proposed in~\cite{karras2018progressive}.
It is a high-quality version of the CelebA dataset consisting of 30,000 images at $1024 \times 1024$ resolution.

\item \textbf{FaceForensics++}:
FaceForensics++  is a forensics dataset containing 1,000 original video sequences that have been manipulated with five automated face manipulation methods: DeepFakes~\cite{DeepFake-url}, Face2Face~\cite{thies2016face2face}, FaceSwap~\cite{FaceSwap-url},
NeuralTextures~\cite{thies2019deferred} and FaceShifter~\cite{li2020advancing}.
FaceForensics++ is a widely-used dataset for evaluating different face manipulation methods.

\end{itemize}

\begin{figure*}[tp]
\centering
\subfigure[]{\includegraphics[width=55mm,height=42mm]{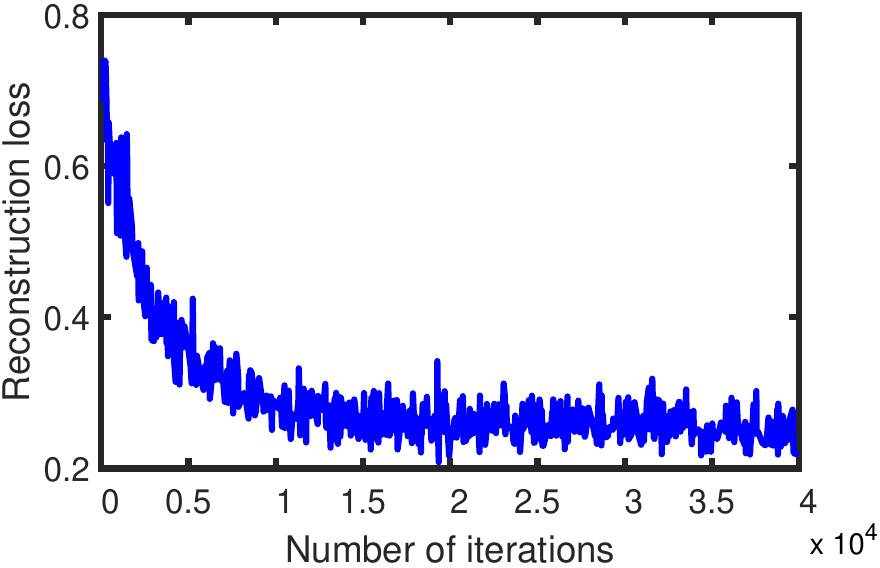}}
\subfigure[]{\includegraphics[width=55mm,height=42mm]{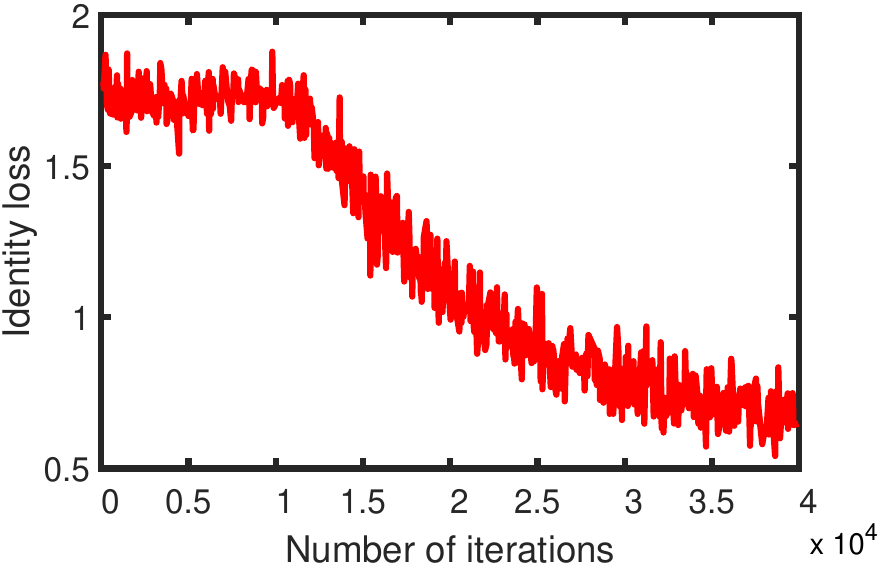}}
\subfigure[]{\includegraphics[width=55mm,height=42mm]{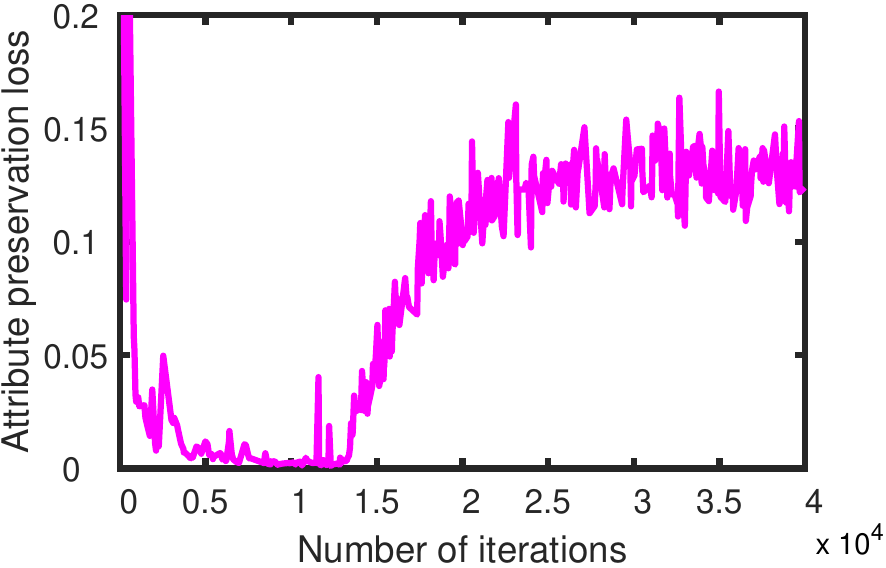}}
\vspace{-4mm}
\caption{Objective functions of our method. (a) The reconstruction loss with respect to different numbers of iterations. (b) The identity loss with respect to different numbers of iterations. (c) The attribute preservation loss with respect to different numbers of iterations. Note that we display only 40,000 iterations for clarity.}
\label{fig:objfunction}
\end{figure*}

Our end-to-end method is implemented based on  Pytorch~\cite{paszke2019pytorch} on a 
Tesla V100 GPU with 32 GB memory.
The discriminator is built with several downsampling residual blocks, which is similar to~\cite{choi2020stargan} except for the multiple-output branch. 
The adaptive moment estimation (Adam) optimizer is used to optimize the network parameters with ${lr} = 0.0001$, ${\beta _1} = 0$, and ${\beta _2} = 0.99$. 
Our method is trained with a batch size of 8, and
the balancing coefficients are set according to an empirical analysis: $\lambda _{id}=10$, $\lambda _{attr}=1$ and $\lambda _{rec}=1$.
The binary masks of all face images are generated by the method proposed in~\cite{yu2018bisenet}.
Note that there are no standard training protocols to train different face swapping algorithms.
Benefiting from the flexible design of our method, we need fewer training samples than previous methods.
In this paper, we randomly choose 30,000 images from the CelebA dataset as the training dataset, and
we test our algorithm on the FaceForensics++ dataset.
Considering that the generalization ability is crutial for face swapping methods,
we also employ the CelebA-HQ dataset as our training set under another experimental setting.
Note that the CelebA-HQ dataset has a different data distribution than the FaceForensics++ dataset, which is mainly composed of high-resolution images with a resolution of $1024 \times 1024$.

Similar to prior methods~\cite{nirkin2019fsgan,li2020advancing}, we adopt three widely
used criteria to test different face swapping algorithms: ID retrieval, pose error and
expression error. 
In addition, Fr$\boldsymbol{\acute{e}}$chet Inception Distance (FID)~\cite{heusel2017gans} is also employed to measure the realism of the generated images.
These evaluation criteria are defined as follows:

\begin{itemize}
\item \textbf{ID retrieval}: ID retrieval measures the identity preservation ability of different face swapping algorithms.
Similar to~\cite{li2020advancing}, we extract the identity information of a face image using the face recognition model proposed by~\cite{wang2018cosface}.
Then, cosine similarity is used to compute the identity difference between the source face image and the target face image. 
Specifically, given a swapped face image, we choose the frame that has the maximum similarity to the given image from the testing set and check whether they belong to the same identity. 
ID retrieval is calculated as the average accuracy of all such retrievals.
\item \textbf{pose error}: Pose error evaluates the pose preservation ability of different face swapping algorithms. 
The head pose is estimated based on the pose estimator proposed by~\cite{doosti2020hope}. 
Then, the ${\ell _2}$ distance of the swapped face image and the target face image is calculated. 
Finally, pose error is obtained as the average error of all pose errors.  
\item \textbf{expression error}: Expression error is utilized to evaluate the expression preservation ability of different face swapping algorithms. 
The expression information of a face image is extracted by a 3D face model~\cite{deng2019accurate}. 
The ${\ell _2}$ distance between the swapped face image and the target face image is computed. 
Expression error describes the average error of all expression errors.
\item \textbf{FID}: FID measures the quality of images by calculating the distance between the distribution of generated images and that of real images. 
Concretely, FID measures the Fr$\boldsymbol{\acute{e}}$chet distance between the distribution of Inception representations~\cite{szegedy2016rethinking} from generated images and real images.
A lower FID score indicates that two distributions are more similar. 
FID is shown to be more stable than previous metrics, such as the structure similarity index method (SSIM) or the Inception score (IS).
\end{itemize}

We evaluate the proposed model against six face swapping methods: FaceSwap~\cite{FaceSwap-url}, DeepFakes~\cite{DeepFake-url}, FSGAN~\cite{nirkin2019fsgan}, IPGAN~\cite{bao2018towards}, FaceShifter~\cite{li2020advancing}, MegaFS~\cite{zhu2021one}, FSLSD~\cite{xu2022high} and RAFSwap~\cite{xu2022region}.
The manipulated sequences of FaceSwap, DeepFakes and FaceShifter are available in the FaceForensics++ dataset, and MegaFS has provided the manipulated sequences on its official website~\cite{MegaFace-url}. 
We select the same sequences for comparison for these methods.
Since the manipulated sequences of FSGAN, FSLSD and RAFSwap are not publicly available,  we carefully re-run the released official source code to generate the manipulated sequences~\cite{FSGAN-url,FSLSD-url,RAFSwap-url} and report the results.
Note that the performance of these methods may differ slightly from the original ones.
The source code of IPGAN is not publicly available, and it is difficult to maintain the original performance if we reimplement this method.
Therefore, we use images provided by~\cite{li2020advancing} for comparisons.
As no official evaluation code exists, we employ the one provided by MegaFS~\cite{zhu2021one} for comparisons.

\subsection{Progressive Face Swapping}
In this subsection, empirical results on the CelebA-HQ dataset are presented to demonstrate the characteristics of our method, where 29,000 images are randomly selected as the training images and 1000 images are used as the testing images.
We first analyze different training loss functions,
and then present some progressive face swapping results.

\flushleft \textbf{Objective functions.} Our method mainly consists of four loss functions:
an identity loss, an attribute preservation loss, a reconstruction loss, and an adversarial loss.
One natural question is the convergence of these loss functions.
Here, we present the convergence curves of different loss functions in Figure~\ref{fig:objfunction} except for the adversarial loss, which describes the objective of a minimax two-player game between the generator and discriminator. 
As shown in Figure~\ref{fig:objfunction}, all loss functions converge to an optimal solution.
From Figure~\ref{fig:objfunction}, we also note another interesting phenomenon.
The attribute preservation loss $\mathcal{L}_{attr}$ decreases rapidly, while the identity loss $\mathcal{L}_{id}$ rises slightly at the beginning. 
Then $\mathcal{L}_{id}$ reaches the highest value, and $\mathcal{L}_{attr}$ obtains the lowest value. We name this process stage I. 
Our method is designed to reconstruct the target image at this stage. 
The attribute preservation loss and the reconstruction loss play a dominant role in synthesizing the face image. However, when the identity loss rises to a certain value, it becomes larger than the attribute preservation and reconstruction losses. 
Our method then synthesizes an image with the same identity as the source image, which will lead to a lower identity loss and a higher attribute preservation loss simultaneously. 
This process is named stage II. 
Finally,  all losses decrease slowly and converge to an optimal solution.

\begin{figure}[tp]
\centering
\includegraphics[width=1.0\linewidth]{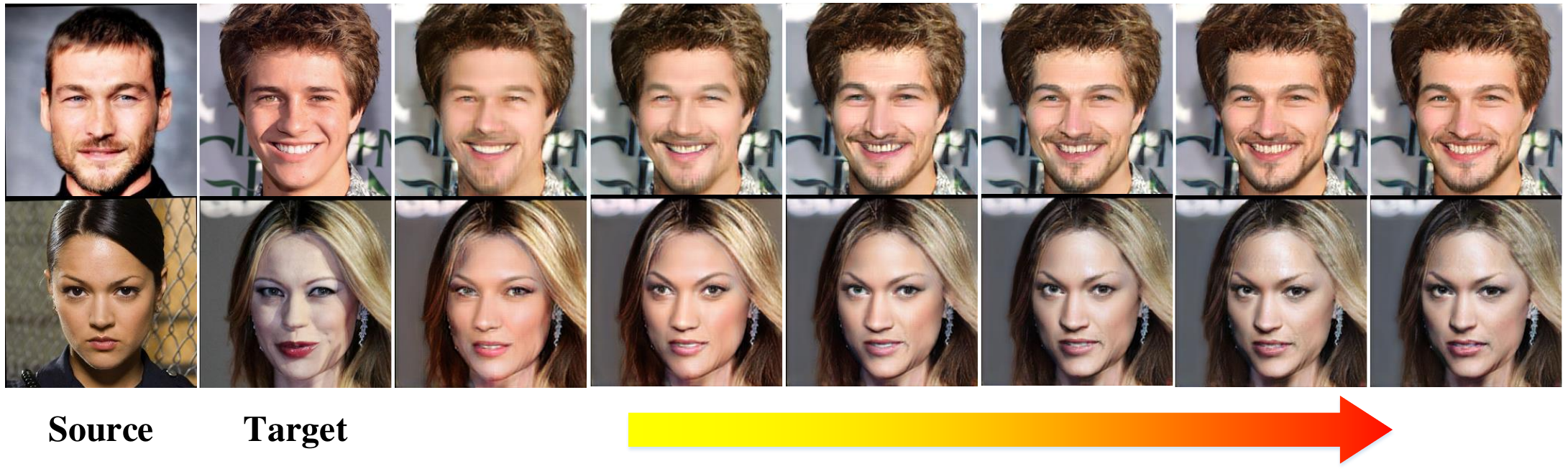}
\vspace{-4mm}
\caption{Progressive face swapping examples with increasing training iterations.}
\label{fig:progressive-fs}
\end{figure}

\flushleft \textbf{Progressive Face Swapping Results.}
Benefiting from the disentanglement of identity and attribute,
our method enables a new face swapping operation: progressive face swapping.
Similar to style interpolation, progressive face swapping refers to
the ability to transfer the source identity to the swapped face image progressively.
Figure~\ref{fig:progressive-fs} shows progressive face swapping examples.
The images in each panel are sorted by increasing training iterations. 
In each row, the leftmost and second-from-the-left images are the source and target images. 
The subsequent
images are the results conditioned on different numbers of training iterations.
Although progressive face swapping results can be obtained by increasing the number of training iterations, the swapped images contain some artifacts due to fewer iterations in training.
Considering the loss function plays a dominant role in this subsection, we change the coefficient of the identity loss, i.e., $\lambda _{id}$, to obtain the progressive face swapping results.

\begin{figure}[tp]
\centering
\includegraphics[width=1.0\linewidth]{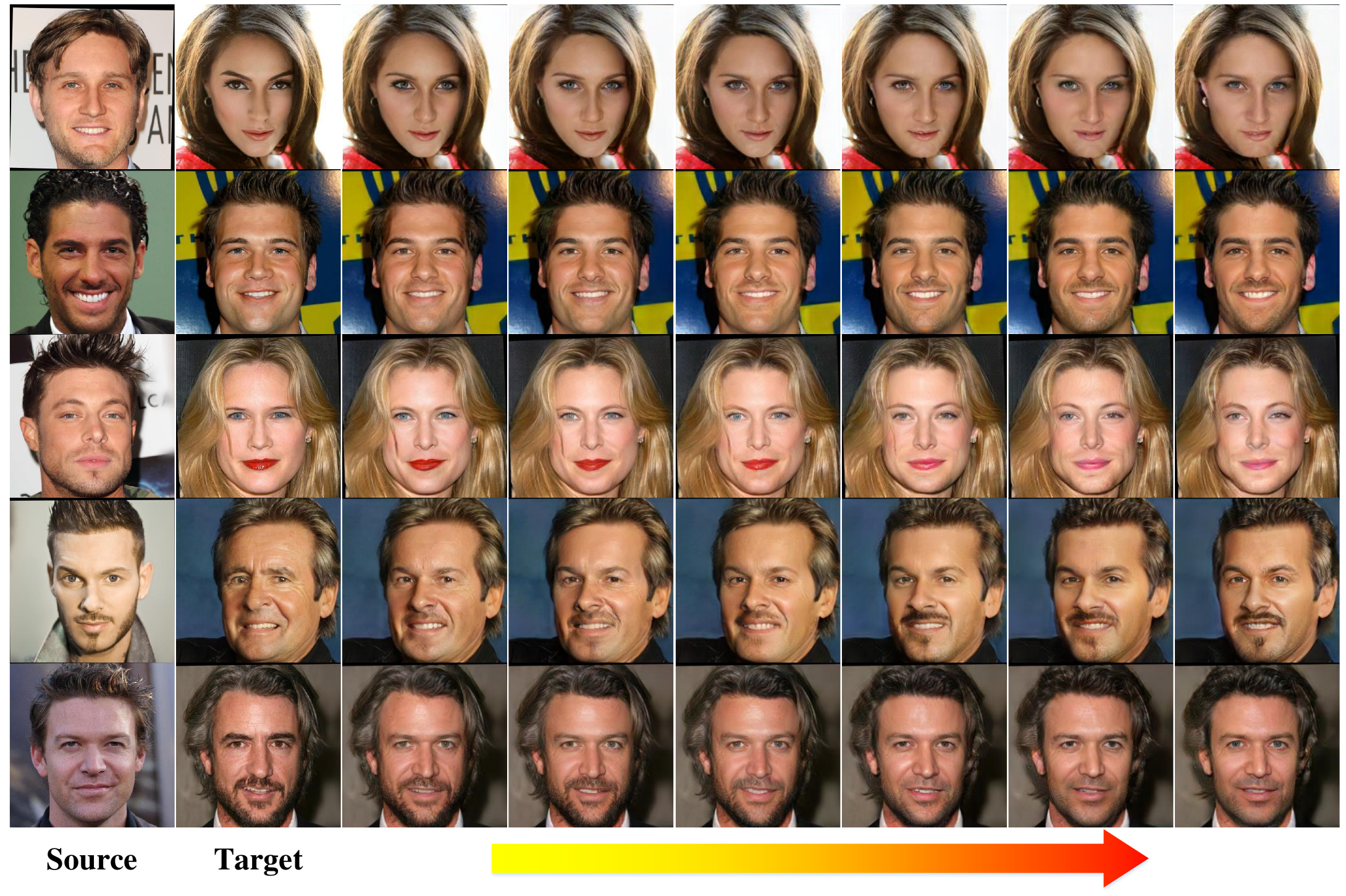}
\vspace{-4mm}
\caption{
Examples of progressive face swapping. The leftmost and second-from-the-left columns represent the source and target images, respectively. The rest columns demonstrate the progressive face swapping results with the increasing coefficients $\lambda _{id}$.}
\label{fig:progressive-id}
\end{figure}

\begin{figure}[tp]
\centering
\includegraphics[width=.8\linewidth]{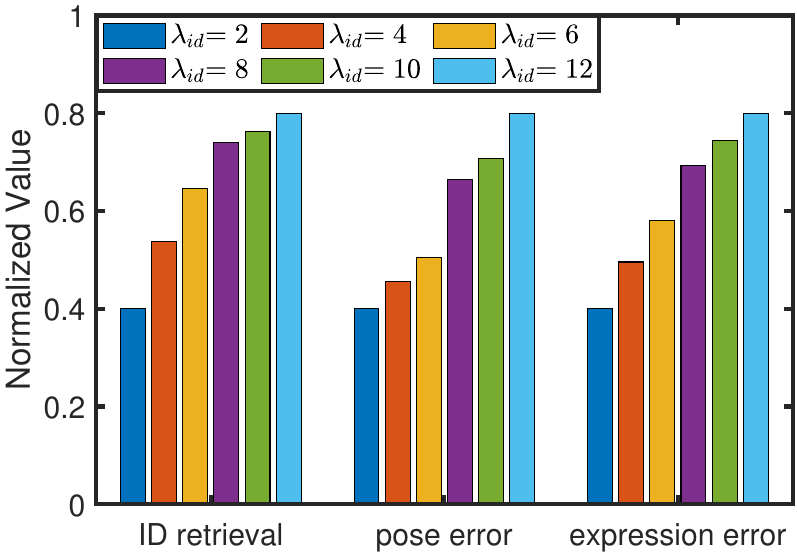}
\vspace{-4mm}
\caption{Evaluation metrics of our method with different coefficients $\lambda _{id}$, which is normalized to 0.4-0.8 for better visualization.}
\label{fig:progressive-value}
\end{figure}

As shown in Figure~\ref{fig:progressive-id}, the generated image is initially similar to the target face image (corresponding to stage I). 
Then, as $\lambda _{id}$ increases, the visual fidelity of the synthetic image increases.
In addition, the identity of the generated image is similar to the identity of the source image (corresponding to stage II). 
However, the similarity of the identity information is low since the facial contour and components still resemble the target face image.
As $\lambda _{id}$ further increases, the identity of the generated image more closely resembles the identity of the source image. 
For example, the facial contour of the generated image is initially similar to that of the target image; then, it changes gradually and is finally similar to the source image. We can also see similar results among the facial components (e.g., eyebrows, noses and mouth) and the complexion.
We have also shown the evaluation metrics of our method with different  $\lambda _{id}$.
Figure~\ref{fig:progressive-value} shows 
the id retrieval increases progressively as $\lambda _{id}$ changes.
Furthermore, the pose error and expression error are also increasing slowly.
When $\lambda _{id}$ increases to a certain level, the increase rate slows down.
The results of the quantitative experiments are consistent with the results of the qualitative experiments.

Note that progressive face swapping has several benefits over traditional face swapping methods.
First, face swapping is a smooth transition between the identity of the source image and the target image in our method. 
We can flexibly steer face swapping to diverse stages, which makes it more useful in real-world scenarios.
In  addition, the identity and attribute information are disentangled more completely, increasing our method's controllability.

\begin{figure}[tp]
\centering
\includegraphics[width=1.0\linewidth]{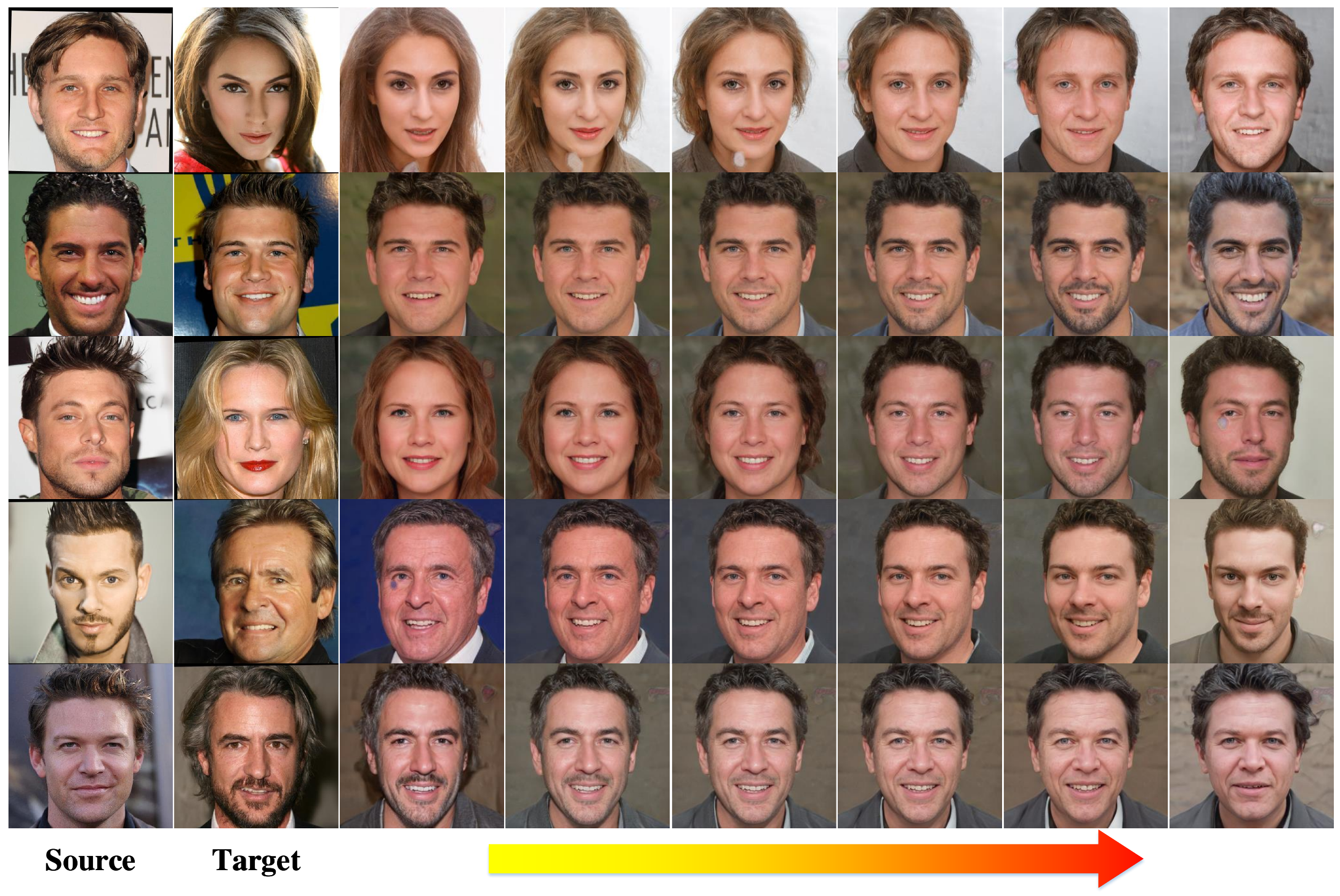}
\vspace{-4mm}
\caption{Examples of disentangled interpolation of identity~\cite{nitzan2020face}. In each line, the identity is extracted from the source and target images, and the attributes are extracted from the target image.  The images generated by the
interpolated values appear in the middle of each of these lines. The interpolated results of our method are already shown in Figure~\ref{fig:progressive-id}.}
\label{fig:progressive-comparison}
\end{figure}

Recent methods~\cite{shen2020interfacegan,shen2021closed} have shown that GANs' latent space is well-behaved and can permit the  editing of facial features via interpolation of the latent code smoothly. 
However, these approaches can only achieve this phenomenon for one-dimensional properties, such as smile or gender.
To the best of our knowledge, FIDLSM~\cite{nitzan2020face} is the only method that can disentangle
face identity from all other face attributes and achieves the smooth editing of these features.
For fair comparisons, the same images in Figure~\ref{fig:progressive-id} are used as the input of FIDLSM.
Figure~\ref{fig:progressive-comparison} demonstrates FIDLSM can interpolate the identity information smoothly and properly.
However, it can hardly preserve the original attribute features of the target image consistently.
For example, in the third row, the identity of the generated face becomes increasingly like the identity of the source image. However, the attributes of the generated face also become
increasingly like the attributes of the source image.
Due to the limited reconstructive ability of the GAN inversion methods, artifacts in the generated face images (e.g., the first row and the third row) exist.

To demonstrate the swapping effect of our method on CelebA-HQ dataset, we also present quantitative results in this paper. Following the approach of MegaFS~\cite{zhu2021one}, we conduct a random swapping of 300,000 pairs of images within CelebA-HQ for testing purposes. The quantitative evaluation metrics encompass pose error, expression error, and FID score. Since ID retrieval between 30,000 original faces and 300,000 swapped faces requires nine billion times of matching, the cosine similarity~\cite{wang2018cosface} of swapped faces and the corresponding source faces is reported to release the computational burden. The comparison results are detailed in Table~\ref{Table:CelebA-HQ-dataset}. It is evident from the table that our method outperforms MegaFace on CelebA-HQ dataset across various quantitative evaluation metrics.

\begin{table}
    \caption{Quantitative experimental results of our method and MegaFS~\cite{zhu2021one} on CelebA-HQ dataset. The quantitative evaluation metrics include cosine similarity, pose error, expression error and FID score.}
    \label{Table:CelebA-HQ-dataset}
    \begin{center}
    \arrayrulewidth=0.6pt
    \begin{tabular}{l|c|c|c|c}
    \hline
    \textbf{Method} & \textbf{\makecell{cosine \\ similarity}  $\uparrow$ } & \textbf{\makecell{pose  \\ error} $\downarrow$}  & \textbf{\makecell{expression \\ error} $\downarrow$} & \textbf{FID $\downarrow$} \\
    \hline
     MegaFS~\cite{zhu2021one}& 0.5014 & 3.58 & 2.87 & 10.16\\
     Ours & \textbf{0.5241} & \textbf{2.96}  & \textbf{2.77} & \textbf{4.05} \\
    \hline
    \end{tabular}
    \end{center}
\end{table}

\subsection{Experimental Results on the FaceForensics++ Dataset}

\begin{figure*}[tp]
\centering
\includegraphics[width=1.0\linewidth]{Img/Celeba_FF/FF++qualitative_comparisons_fs_paper.pdf}
\vspace{-4mm}
\caption{Evaluation results of different face swapping methods on the FaceForensics++ dataset. From left to right, the columns show images of the source, the target, and the results of FaceSwap, DeepFakes, FSGAN, IPGAN, FaceShifter, MegaFS and our method. Note that IPGAN images are provided by~\cite{li2020advancing}.}
\label{fig:FF-qualitative-comparisons-fs-paper}
\end{figure*}

\textbf{Data Cleaning.} The FaceForensics++ dataset contains 1,000 pristine video sequences. 
It is a benchmark dataset used to test different face swapping methods.
The synthetic results of three face swapping methods (i.e., Deepfakes, FaceSwap and FaceShifter) and two face reenactment methods (i.e., Face2Face and NeuralTextures) are available for this dataset.
We reorganize this dataset to  evaluate face manipulation methods. %
First, we sample 10 frames evenly from each video sequence, and 10,000 frames are selected. 
The multitask cascaded convolutional neural network (MTCNN)~\cite{zhang2016joint} is then employed to detect and align the faces in these frames. 
We manually check the face images: if there is no face or the faces are incorrectly detected in the frame, we choose another nearby frame that contains the correct face image. Finally, 10,000 face images are obtained.
For the provided videos of different face manipulation methods, we sample frames with the same index as the original video sequences, and we detect and align face images in the manipulated frames for comparisons.
The pristine video sequences in the FaceForensics++ dataset are downloaded from the internet,
and many videos may belong to the same identity.
For example, videos \#043 and \#343 show the same person, Vladimir Putin, and videos \#179, \#183 and \#826 contain the same person, Barack Obama.
To enable a fair comparison in terms of ID retrieval, accurate identity labels for all video sequences are needed. 
Therefore, we extract the identity vectors of all face images and cluster these identity vectors automatically. Moreover, we manually check the correctness of the identity labels. Ultimately, we obtain 885 identities for all face images.

\flushleft \textbf{Qualitative Evaluation.}
The qualitative evaluations of manipulation methods are shown in Figure~\ref{fig:FF-qualitative-comparisons-fs-paper}, which
displays source and target images with large expression, pose, illumination and age variations. 
Benefiting from the input of video pairs (i.e., source and target video pairs), FaceSwap and DeepFakes successfully swap the identity of the source and target images. 
However, the identity and attribute information are not well disentangled. 
For example, the swapped image has similar illumination to the source image in row 3. In addition, the swapped images show noticeable artifacts in rows 1-5  because of their simple training and blending strategy.
While IPGAN achieves better swapping results than FaceSwap and DeepFakes, it still suffers from a decreased image resolution and fails to preserve face contours with the target faces (e.g., rows 1-5),  which may reduce the identity similarity between the swapped faces and source faces.
The results of FSGAN are much more pleasing, as they appear more natural and contain no apparent artifacts.
However, the disentanglement of FSGAN is not as good in two aspects: 
(1) The identity information of the swapped images is quite similar to that of the target images. (2) The illumination of the swapped images fails to preserve that of the target images.

\begin{table*}
\caption{Quantitative results of on the FaceForensics++ dataset. 
Results of FaceSwap~\cite{FaceSwap-url}, DeepFakes~\cite{DeepFake-url}, FaceShifter~\cite{li2020advancing} and MegaFS~\cite{zhu2021one} are obtained from their websites.
Results of FSGAN~\cite{nirkin2019fsgan}, FSLSD~\cite{xu2022high}, and RAFSwap~\cite{xu2022region} are evaluated based on the released code and models.}
\label{Table:firstexp-ff++}
\begin{center}
\arrayrulewidth=0.6pt
\vspace{-3mm}
\begin{tabular}{l|c|ccccc}
\hline
\textbf{Category} & \textbf{Method} & \textbf{ID retrieval $\uparrow$ } & \textbf{pose error $\downarrow$} & \textbf{expression error $\downarrow$} & \textbf{FID $\downarrow$} \\
\hline
\multirow{2}{*}{face reenactment} &Face2Face~\cite{thies2016face2face} & - & 2.68 & 2.09 & 3.52 \\
& NeuralTextures~\cite{thies2019deferred} &  - & 2.21 & 1.64 & 2.58 \\
\hline
\makecell{face reenactment \\  \& face swapping }&FSGAN~\cite{nirkin2019fsgan} & 21.33 & 2.04 & 1.98 & 5.99 \\
\hline
\multirow{7}{*}{face swapping} &FaceSwap~\cite{FaceSwap-url}& 72.69 & 2.58 & 2.89 & 4.07\\
& DeepFakes~\cite{DeepFake-url} &  88.39 & 4.64 & 3.33 & 4.64  \\
& FaceShifter~\cite{li2020advancing}&  90.68 & 2.55 & 2.82 & 4.53  \\
& MegaFS~\cite{zhu2021one}&   90.83 & 2.64 & 2.96 & 11.04  \\
& FSLSD~\cite{xu2022high}&  90.05 & 2.46 & 2.79 & 25.99 \\
& RAFSwap~\cite{xu2022region}&   92.54 &3.21 & 3.60 & 26.61 \\
& Ours & \textbf{94.48} & \textbf{2.10} & \textbf{2.69} & \textbf{4.21} \\
\hline
\end{tabular}
\end{center}
\end{table*}

Since MegaFS, FSLSD and RAFSwap are mainly designed for high-resolution face swapping based on the pre-trained StyleGAN network, large discrepancies in skin colors and
illuminations between the target images and the generated images  can be seen in these methods (e.g., row 2 and row 4 of RAFSwap). 
In addition, MegaFs and FSLSD generate distinct face contours and ignore the source face shape because of the explicit blending of the generated images with the background.
There are noticeable artifacts in the swapped results for these two methods (e.g., rows 2-3).
Among all the methods, FaceShifter and our method generate visually appealing face swapping results.
However, FaceShifter cannot completely disentangle identity and attribute information because of its fixed identity encoder.
In contrast, benefiting from the flexible framework,
our method can generate more realistic face images that can faithfully preserve the identity of the source images while having the same attribute information as the target images. 
Taking the synthetic images of our method in rows 1-5 as examples, the nose in row 1 and eyes in row 2 are similar to those of the source images, the eyebrow and beard are more similar to the source image in row 4, and the eye and mouth are more natural and realistic in row 5.

\flushleft \textbf{Quantitative Evaluation.}

As face swapping aims to
aesthetically synthesize an image with 
the identity of the source image and the attributes of the target image simultaneously.
Thus, more comprehensive evaluations are provided with quantitative analysis.
The evaluation results of different methods are shown in Table~\ref{Table:firstexp-ff++}.
Note that Face2Face and Neural Textures are face reenactment methods, so we do not
report their ID retrieval values in this table.

ID retrieval in Table~\ref{Table:firstexp-ff++} shows the identity preservation ability of the evaluated methods. 
FaceSwap and DeepFakes achieve relatively low ID retrieval scores.
This can be attributed to that FaceSwap is essentially a graphics-based approach, and DeepFakes can only exploit a few identities to learn the swapping model.
FSGAN almost fails to preserve the identity information of the source images, and its ID retrieval is only 21.33\%. 
Note that FSGAN is a framework designed for both face swapping and face reenactment. 
The face contour of the generated images in FSGAN remains the same as that of the target images, and the loss functions for face swapping in FSGAN are oversimplified. 
As such, FSGAN has lower pose and expression errors than other face swapping methods. 
FaceShifter achieves a higher ID retrieval than other methods, benefiting from the identity integration module.
MegaFS and FSLSD perform comparably to that of FaceShifter in terms of ID retrieval.
Benefiting from the sophisticated design of the global and local feature interaction module, RAFSwap 
achieves a higher retrieval performance than MegaFS and FSLSD.  
Our method also performs well against RAFSwap in these experiments. 
The reason is that the identity encoder of our method is learned automatically during the training process, which is much more flexible than the fixed identity encoder used by FaceShifter or the feature interaction module of RAFSwap.

Pose error and expression error in Table~\ref{Table:firstexp-ff++} show the attribute preservation ability of the evaluated face reenactment and swapping methods.
Face reenactment methods like NeuralTextures and Face2Face usually have lower pose and expression errors than face swapping methods. 
The reason is that face reenactment methods are mainly designed to control facial movements and expression deformations while neglecting the swapping of the identity information. 
Regarding the face swapping methods, FaceShifter and FSLSD perform better than the other methods except for ours. 
Our method performs favorably against FaceShifter and FSLSD in terms of pose error. 
The pose error of our method is 2.10, while the pose error of FaceShifter and FSLSD is 2.55 and 2.46, respectively. 
The expression error of our method is 2.69, while the expression error of FaceShifter and FSLSD is 2.82 and 2.79, respectively.
Our method performs well in identity preservation and comparably to face reenactment methods in preserving pose and expression. 
The identity and attribute preservation abilities are, to a certain degree, contradictory to each other.
As an extreme case, if we use all of the target images as the swapped results, then ID retrieval should be close to zero (the method should fail to maintain the identity of the source images). 
However, the pose and expression errors are also close to zero in this case (the method faithfully preserves the attributes of the target images). 
Overall, our method strikes a balance between
the identity preservation ability and the attribute preservation ability. It performs the best among all of the face swapping methods.

The last column of Table~\ref{Table:firstexp-ff++} shows FIDs of different comparison methods. 
From the table, we can get the following conclusions.
First, face reenactment methods usually have lower FID scores than face swapping methods, as they warp the texture directly from the images.  
Second, StyleGAN-based methods, i.e., MegaFS, FSLSD, RAFSwap, usually have higher FID scores than other methods. 
The reason is that the images generated by StyleGAN have a quite different distribution from the images in the FaceForensics++ dataset. 
Third, FaceSwap achieves the lowest FID score among all face swapping methods, because it also directly warps the texture from the source images.   
Our method achieves a comparable  FID score with FaceSwap and performs favorably against all the other face swapping methods. The results further validate that the quality of our generated images is better than recent state-of-the-art face swapping methods.

\subsection{Results of Challenging Cases}

\begin{figure*}[tp]
\centering
\includegraphics[width=0.90\linewidth]{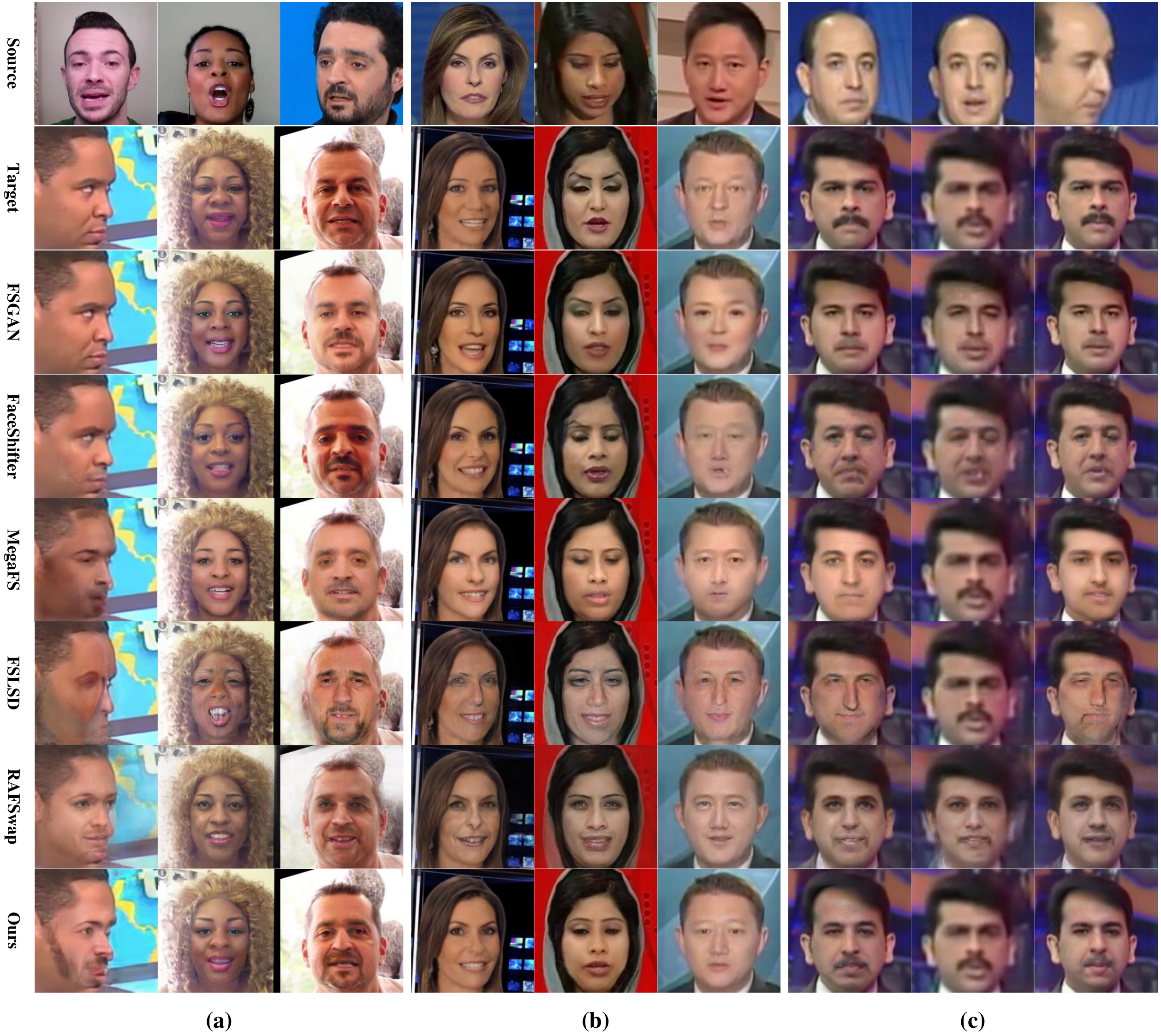}
\vspace{-4mm}
\caption{Examples of our method with the largest pose error, the largest expression error and the minimum identity similarity on FaceForensics++ dataset. (a) illustrates first three results with the largest pose error. (b) displays first three results with the largest expression error. (c) denotes first three results with the minimum identity similarity. From the top to the bottom rows are images of the source, the target, and the results of FSGAN, FaceShifter, MegaFS, FSLSD, RAFSwap and our method (zoomed in for a better view).}
\label{fig:FF++qualitative_comparison_maxerror}
\end{figure*}

\begin{figure*}[tp]
\centering
\includegraphics[width=0.9\linewidth]{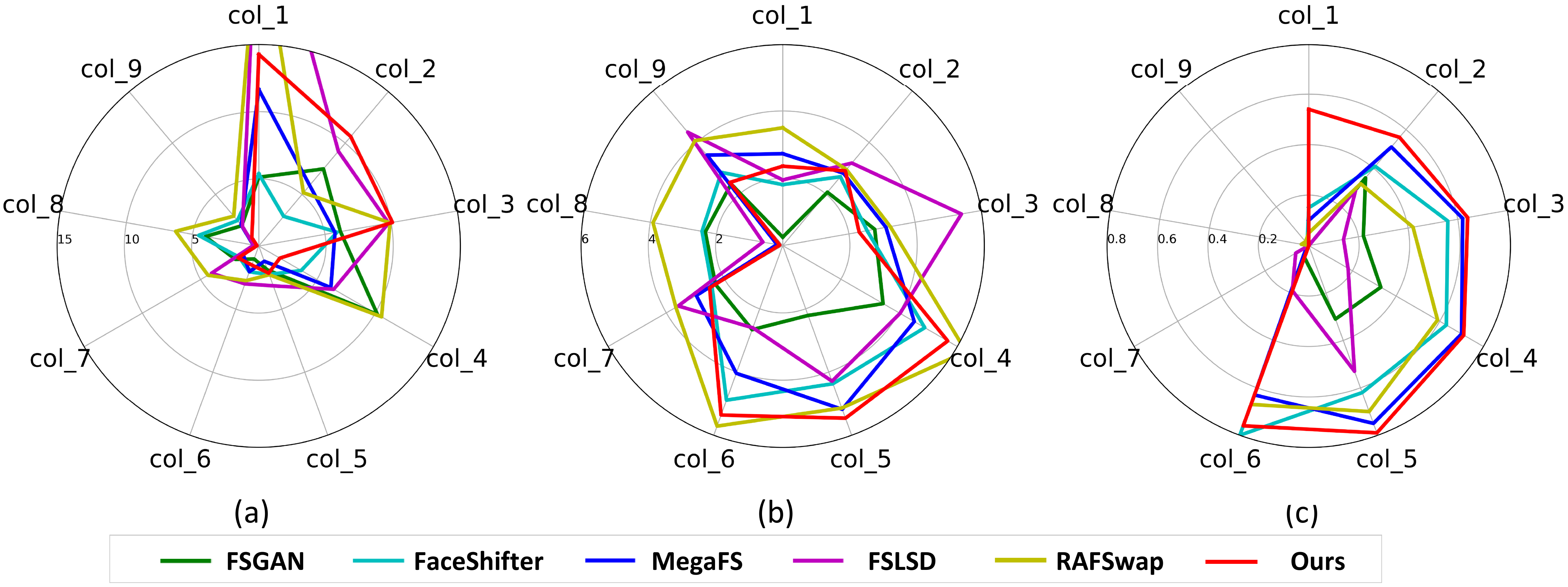}
\vspace{-4mm}
\caption{Ranking of quantitative metrics for different methods. (a) Pose error ranking; (b) Expression error ranking; (c) Identity similarity ranking.}
\label{fig:FF++qualitative_comparison_maxerror2}
\end{figure*}

To evaluate the limitations of our method, we show several examples with the largest pose error, expression error, and minimum identity similarity.
Figure~\ref{fig:FF++qualitative_comparison_maxerror} illustrates the most challenging results, which we thoroughly analyze in this section. 
Specifically, Figure~\ref{fig:FF++qualitative_comparison_maxerror}(a) and Figure~\ref{fig:FF++qualitative_comparison_maxerror}(b) display the first three results with the largest pose error and expression error, respectively. 
While Figure~\ref{fig:FF++qualitative_comparison_maxerror}(c) shows the first three results with the minimum identity similarity. 
We also present the results of FSGAN, FaceShifter, MegaFS, FSLSD, and RAFSwap for comparison. 
As expected, both methods encounter significant challenges due to large pose, expression, and resolution variations between source and target images. 
Taking the first column of Figure~\ref{fig:FF++qualitative_comparison_maxerror}(a) as an example, the target image exhibits an extreme pose (near ${90^ \circ }$). 
The identity of the source image and the attributes of the target image are well preserved in our results. 
However, there are some inconsistencies caused by the inaccurate mask of the target image and the existence of the beard in the source image. 
A similar phenomenon can be seen in the results of MegaFS, FSLSD and RAFSwap.   
In contrast, there is a subtle difference between the target image and the swapped result synthesized by FaceShifter.
However, the beard of the source image is not preserved well in the swapped image for FaceShifter. 
Additionally, FSGAN struggles to preserve identity in the swapped images.
As another example, the face images in the second column of Figure~\ref{fig:FF++qualitative_comparison_maxerror}(b) contain large discrepancies in attributes. Despite this, our method successfully preserves the identity of the source image (e.g., the face contour and forehead) and the attributes of the target image (e.g., eyebrow and mouth).
Nearly all of the compared methods fail to generate the swapped image.
For instance, FaceShifter fails to fully disentangle identity and attribute information, resulting in confused eyebrows.
Offering another illustration, the face images in the second column of Figure~\ref{fig:FF++qualitative_comparison_maxerror}(c) exhibit substantial resolution differences and significant disparities in facial contour.  Most methods struggle to produce an ideal face swapping result. For example, artifacts are present in the swapped images for both FaceShifter and RAFSwap. While MegaFS, FSLSD, and our method effectively preserve pose and expression, the identity similarity between the source and swapped images remains relatively low across almost all methods.

Suppose that from left to right, the images in each column of Figure~\ref{fig:FF++qualitative_comparison_maxerror} are denoted as col\_1, ..., col\_9, respectively. 
For a more thorough analysis of challenging cases, we show the quantitative metric ranking of various methods in Figure~\ref{fig:FF++qualitative_comparison_maxerror2}. 
Since there are only 14 identities in this figure, we also report the cosine similarity of swapped images and the corresponding source
images as identity similarity to mitigate random interference.
Figure~\ref{fig:FF++qualitative_comparison_maxerror2} displays rankings for pose error, expression error, and identity similarity across different methods.
From the figure, we can get the following conclusions. 
Firstly, our method performs slightly worse than FaceShifter and is comparable with RAFswap regarding pose error ranking and expression error ranking. 
However, our method achieves the best performance in terms of identity similarity. 
Taking the results of col3 as an example,  the pose error of our method is comparable with other methods.
For the expression error and identity similarity, our method performs favorably against all the other methods.
Secondly, our method performs comparably to others regarding identity similarity in col7, col8, and col9, yet outperforms them in pose and expression error rankings. 
It is worth noting that although Figure~\ref{fig:FF++qualitative_comparison_maxerror} and Figure~\ref{fig:FF++qualitative_comparison_maxerror2} highlight challenging cases for our method, the synthesized results remain competitive with other approaches.

\begin{figure*}[tp]
\centering
\includegraphics[width=0.90\linewidth]{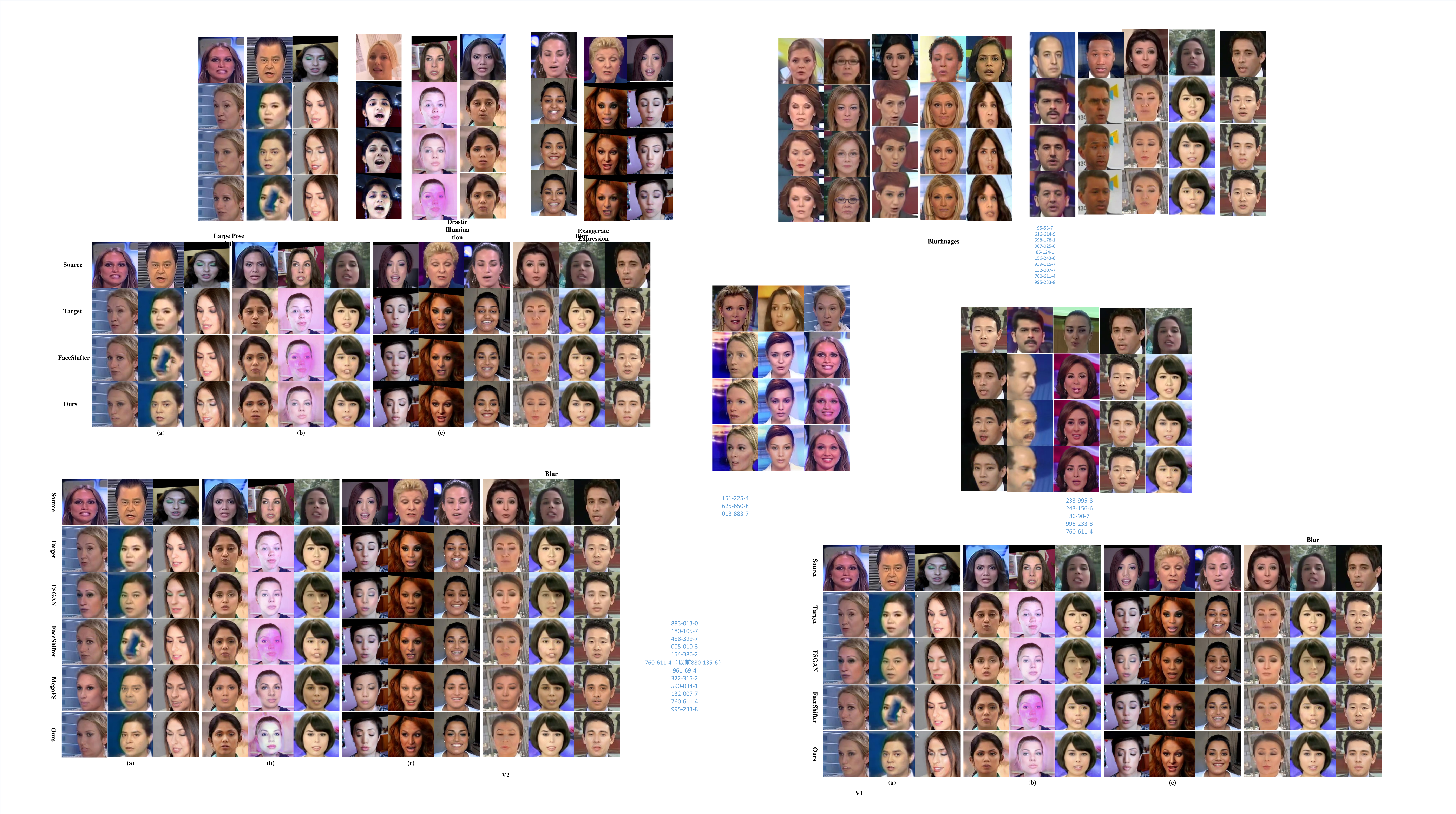}
\vspace{-4mm}
\caption{Results of FSGAN, FaceShifter, MegaFS, FSLSD, RAFSwap and our method on the FaceForensics++ dataset with different challenges: (a) large pose variations, (b) drastic lighting conditions, and (c) exaggerated expressions (magnified for clarity).}
\label{fig:FF++qualitative-comparison-more}
\end{figure*}

Additional results of FSGAN, FaceShifter, MegaFS, FSLSD, RAFSwap and our method on the FaceForensics++ dataset with different challenges are shown in Figure~\ref{fig:FF++qualitative-comparison-more}.
Figure~\ref{fig:FF++qualitative-comparison-more}(a) shows source and target images with large pose variations. 
Our method can synthesize visually convincing results while faithfully preserving the pose information.
The semantic information provides a strong prior for our method, which is carefully designed to avoid the effect of complex backgrounds and to guide our approach to generate a satisfactory result.
Figure~\ref{fig:FF++qualitative-comparison-more}(b) shows source and target images with drastic lighting conditions.
FSGAN, FaceShifter, MegaFS and FSLSD cannot preserve the lighting conditions of the target images well. 
Furthermore, obvious artifacts can be seen in the results of FaceShifter. 
The attributes are not preserved well in the generated images for RAFSwap (e.g., the eye status). In summary, under extremely unnatural lighting conditions, all other face swapping methods fail to synthesize visually appealing results. 
Benefiting from the disentangled representation module and the semantic-guided fusion module, our method can successfully separate the identity and attribute information.
Both FSGAN and our method can maintain the expression of the target images. 
However, FSGAN fails to keep the identity of the source images. 
Several factors contribute to the performance of our model.
First, the identity and attribute information is disentangled in our method. 
Moreover, landmark features are leveraged to transmit the expression and pose information of the target image into the semantic-guided fusion module to learn the pose and expression information precisely. 
In addition, the attribute preservation loss is adopted to ensure that the synthetic and target images are similar in the multilevel feature space.

\begin{table}
\caption{User study results on the FaceForensics++ dataset. The numerical value indicates the percentage of users who select this algorithm.}
\label{Table:user-study-ff++}
\begin{center}
\arrayrulewidth=0.6pt
\vspace{-3mm}
\begin{tabular}{l|c|c|c}
\hline
\textbf{Method} & \textbf{\makecell{identity \\ preservation}  $\uparrow$ } & \textbf{\makecell{attribute \\ preservation} $\uparrow$}  & \textbf{Quality $\uparrow$} \\
\hline
FaceSwap~\cite{FaceSwap-url}& 15.65 & 12.46 & 5.55\\
DeepFakes~\cite{DeepFake-url} & 17.53 & 5.25  & 2.77  \\
FaceShifter~\cite{li2020advancing}& 18.39 & 23.23  & 17.86 \\
MegaFS~\cite{zhu2021one}& 12.02 & 7.13 & 12.36  \\
FSLSD~\cite{xu2022high}& 1.46 & 1.14 & 1.83 \\
RAFSwap~\cite{xu2022region}&   9.03 & 4.32& 14.62 \\
Ours & \textbf{25.92} & \textbf{46.47} & \textbf{45.04} \\
\hline
\end{tabular}
\end{center}
\end{table}

\subsection{User Evaluation}%

In this section,  a user study is conducted to evaluate the performance of each method on the FaceForensics++ dataset.
In order to evaluate the performance and realism of different methods, 200 pairs of images
from the FaceForensics++ dataset are given to 50 users to answer the following questions. 1) Which one has the most similarity identity with the source face? 2) Which one better preserves the attributes (i.e., pose and expression) of the target face? 
3) Which one is the most realistic result? 
The answers from 50 users are reported in Table~\ref{Table:user-study-ff++}. 
As reported in Table~\ref{Table:user-study-ff++}, our method achieves the highest score for identity similarity, attribute preservation ability, and the realism of the results.

\subsection{Ablation Study}
\begin{figure}[tp]
\centering
\includegraphics[width=1.0\linewidth]{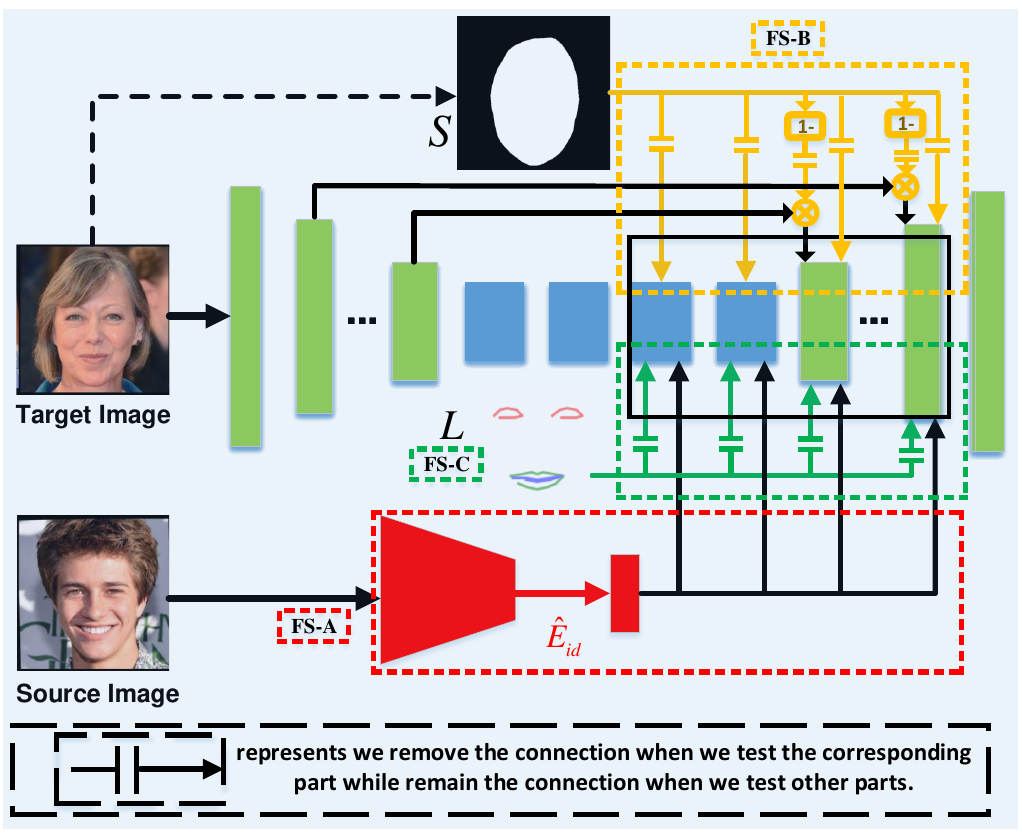}
\vspace{-4mm}
\caption{Illustration of three variants of our method. FS-A represents that the identity encoder ${E_{id}}$ is replaced with a pre-trained face recognizer ${\hat E_{id}}$. FS-B denotes that we remove the semantic mask information from our method. FS-C means that the landmark information is eliminated from our method. Note that there are also modifications in SDL for FS-A and FS-B, which we omit for convenience.}
\label{fig:ablation-network-structure}
\end{figure}

We conduct an ablation study on the FaceForensics++ dataset to analyze the contributions of the individual components of our method.
Three variants of our method are designed as follows.
First, we replace our identity encoder with a fixed identity vector extracted from a pre-trained face recognizer, e.g., Arcface~\cite{deng2019arcface}, and leave all the other parts intact. 
We refer to this method as FS-A. 
FS-A is implemented to evaluate the effectiveness of DRM.
Second, we slightly change the structure of SDL and remove the semantic mask information in SDL. 
We denote this approach as FS-B.
Finally, we eliminate the landmark information in SDL and leave all the other parts intact; this scheme is refered to as FS-C. 
FS-B and FS-C are designed to verify the effectiveness of SFM.
The differences between the three variants of our method and the original method are shown in Figure~\ref{fig:ablation-network-structure}. 
The modification of FS-A is depicted in a red dotted box, while the modifications of FS-B and FS-C are marked with an orange dotted box and a green dotted box, respectively.

\begin{figure}[tp]
\centering
\includegraphics[width=1.0\linewidth]{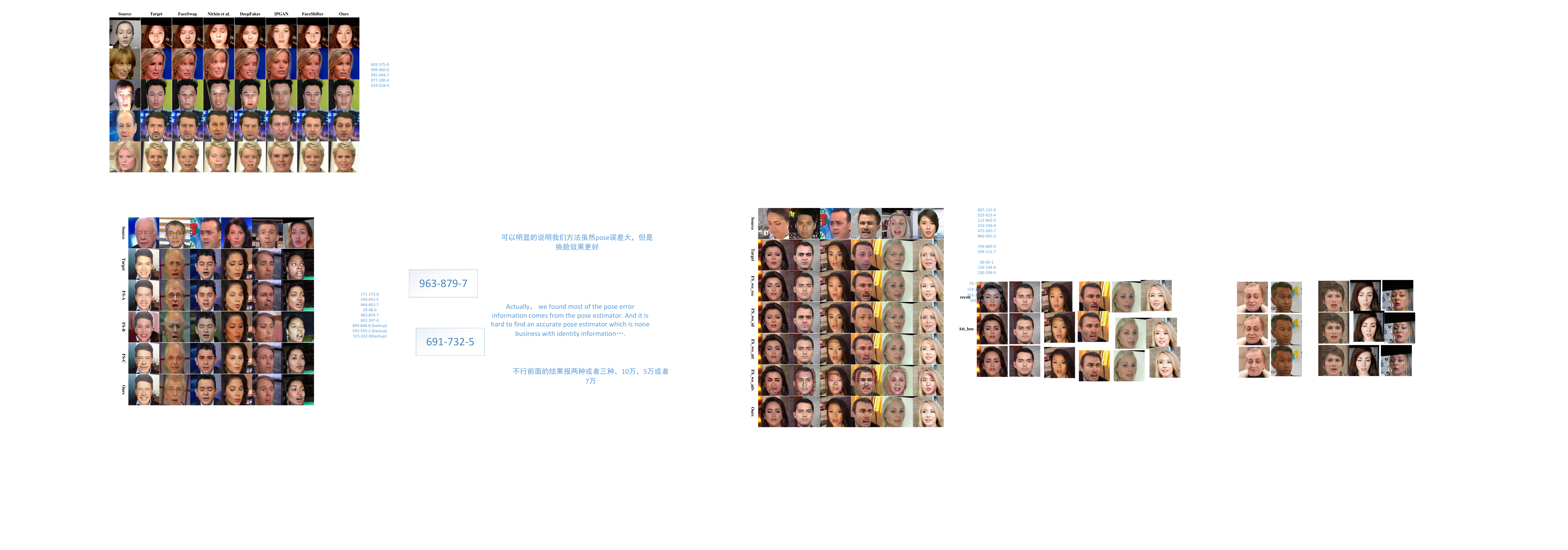}
\vspace{-4mm}
\caption{Qualitative results of different variants of our method on the FaceForensics++ dataset. }
\label{fig:ablation-study}
\end{figure}

Figure~\ref{fig:ablation-study} shows the evaluation results of all the variants.  
FS-A can swap the identity of the source and target images.
However, the detailed identity information is missing. 
There are also some artifacts in the synthesized results.
Taking the first two columns of FS-A as an example, the wrinkles
disappear in the first column, and obvious artifacts can be seen in the second column.
FS-B cannot preserve the identity of the source images well. 
The synthesized results seem to combine the source and target images, and we can hardly see the swapping effect.
These results demonstrate that semantic information is essential for our method if we have limited training samples.
The generated results of FS-C fail to maintain the attribute information of the target images (e.g., pose and expression information).
Table~\ref{Table:ablation-study-ff++} shows the quantitative results.
Note that identity preservation performance decreases significantly if we remove the semantic mask information, which means FS-B cannot successfully swap the identities of the source and target images. 
However, FS-B faithfully maintains the expression information of the target images.

\begin{table}
\caption{Quantitative results of the variants of our method on the FaceForensics++ dataset.}
\label{Table:ablation-study-ff++}
\begin{center}
\arrayrulewidth=0.6pt
\begin{tabular}{l|ccc}
\hline
\textbf{Method} & \textbf{ID retrieval $\uparrow$ } & \textbf{pose error $\downarrow$}  & \textbf{expression error $\downarrow$}\\
\hline
FS-A & 95.56 & 3.09 & 3.22 \\
FS-B & 1.20 & 3.83 & 2.63 \\
FS-C & 93.07 & 3.75 & 3.50 \\
Ours &94.48 & 2.10 & 2.69 \\
\hline
\end{tabular}
\end{center}
\end{table}

\begin{figure}[tp]
\centering
\includegraphics[width=1.0\linewidth]{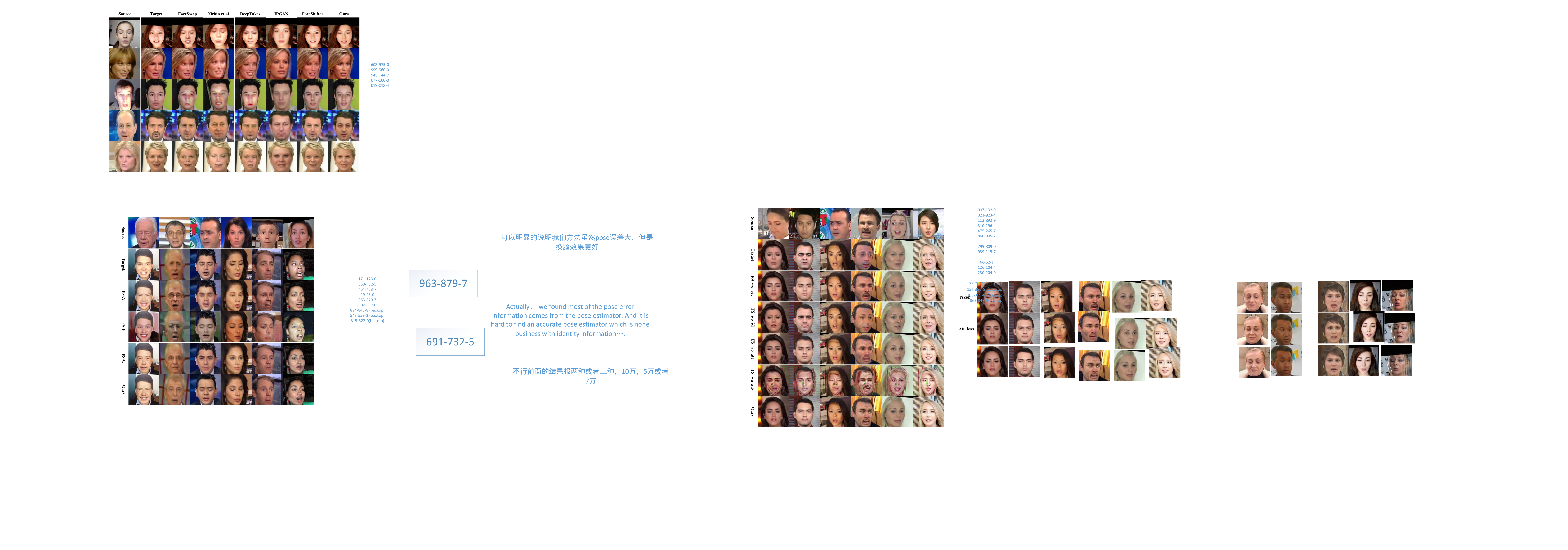}
\vspace{-4mm}
\caption{Qualitative results of the effect of different loss functions on the FaceForensics++ dataset. }
\label{fig:ablation-study-loss}
\end{figure}

We analyze the effect of loss functions in the proposed model. 
by excluding the reconstruction loss (FS\_wo\_rec),   identity preservation loss (FS\_wo\_id), attribute preservation loss (FS\_wo\_attr), and adversarial loss (FS\_wo\_adv), respectively. 
The qualitative and quantitative results trained with different loss configurations are shown in Figure~\ref{fig:ablation-study-loss} and Table~\ref{Table:ablation-study-ff++loss}. 
These results show that it is not feasible to generate a realistic swapped image without the adversarial and identity losses. 
Our method can generate a swapped image without the reconstruction and attribute preservation losses. 
However, the results are far from satisfactory.
Configurations with one of these losses being disabled lead to degenerated visual results, which shows that these loss functions are crucial for our method.

\begin{table}
\caption{Effect of loss Functions on the FaceForensics++ dataset.}
\label{Table:ablation-study-ff++loss}
\begin{center}
\arrayrulewidth=0.6pt
\begin{tabular}{l|ccc}
\hline
\textbf{Method} & \textbf{ID retrieval $\uparrow$ } & \textbf{pose error $\downarrow$}  & \textbf{expression error $\downarrow$} \\
\hline
FS\_wo\_rec & 90.14 & 2.67 & 2.99 \\

FS\_wo\_id & 0.24 & 1.80 & 2.01 \\

FS\_wo\_att & 89.30 & 2.68 & 2.97 \\

FS\_wo\_adv & 92.81 & 7.84 & 3.72 \\

Ours &94.48 & 2.10 & 2.69 \\
\hline
\end{tabular}
\end{center}
\end{table}

\section{Conclusion and Future Work}
\label{sec:conclusions}
This paper presents an end-to-end framework for one-shot progressive face swapping based on Generative Adversarial Networks.
Our method consists of a disentangled representation module and a semantic-guided fusion module. 
The disentangled representation module comprises an attribute encoder, an identity encoder, and a semantic-guided fusion decoder. 
The identity encoder is more flexible, and the attribute encoder contains more details of the attributes than existing face swapping methods.
Semantic masks and facial landmarks are further introduced into the semantic-guided fusion module to control the swapped area and more accurately model the pose and expression. 
Benefiting from the flexible design of our method, we achieve state-of-the-art performance on various benchmark datasets with fewer training samples.

Although the proposed approach achieves more accurate and photorealistic swapping results than other methods, some future studies still need to be considered.
First, completely disentangling the identity and attribute information in extreme cases is still a challenging problem. Large pose variations, drastic lighting conditions, and exaggerated expressions pose significant challenges for face swapping.
In the future, we will focus on solving this problem so that we can control and edit face images more accurately. In addition, a solution to face swapping is provided, and face reenactment is given less attention in this paper. 
Like face swapping, one of the challenging problems for face reenactment is the disentanglement of the attribute information. A promising direction of future research is to design a disentangled framework for face swapping and face reenactment simultaneously.

\ifCLASSOPTIONcaptionsoff
  \newpage
\fi
\bibliographystyle{IEEEtran}
\bibliography{mytrans}

\begin{thebibliography}{10}
\providecommand{\url}[1]{#1}
\csname url@samestyle\endcsname
\providecommand{\newblock}{\relax}
\providecommand{\bibinfo}[2]{#2}
\providecommand{\BIBentrySTDinterwordspacing}{\spaceskip=0pt\relax}
\providecommand{\BIBentryALTinterwordstretchfactor}{4}
\providecommand{\BIBentryALTinterwordspacing}{\spaceskip=\fontdimen2\font plus
\BIBentryALTinterwordstretchfactor\fontdimen3\font minus
  \fontdimen4\font\relax}
\providecommand{\BIBforeignlanguage}[2]{{%
\expandafter\ifx\csname l@#1\endcsname\relax
\typeout{** WARNING: IEEEtran.bst: No hyphenation pattern has been}%
\typeout{** loaded for the language `#1'. Using the pattern for}%
\typeout{** the default language instead.}%
\else
\language=\csname l@#1\endcsname
\fi
#2}}
\providecommand{\BIBdecl}{\relax}
\BIBdecl

\bibitem{blanz1999morphable}
V.~Blanz and T.~Vetter, ``A morphable model for the synthesis of 3d faces,'' in
  \emph{Proceedings of the Annual Conference on Computer Graphics and
  Interactive Techniques}, 1999, pp. 187--194.

\bibitem{tran2018nonlinear}
L.~Tran and X.~Liu, ``Nonlinear 3d face morphable model,'' in \emph{Proceedings
  of the IEEE Conference on Computer Vision and Pattern Recognition}, 2018, pp.
  7346--7355.

\bibitem{goodfellow2014generative}
I.~Goodfellow, J.~Pouget-Abadie, M.~Mirza, B.~Xu, D.~Warde-Farley, S.~Ozair,
  A.~Courville, and Y.~Bengio, ``Generative adversarial nets,'' in
  \emph{Advances in Neural Information Processing Systems}, 2014, pp.
  2672--2680.

\bibitem{karras2019style}
T.~Karras, S.~Laine, and T.~Aila, ``A style-based generator architecture for
  generative adversarial networks,'' in \emph{Proceedings of the IEEE
  Conference on Computer Vision and Pattern Recognition}, 2019, pp. 4401--4410.

\bibitem{karras2020analyzing}
T.~Karras, S.~Laine, M.~Aittala, J.~Hellsten, J.~Lehtinen, and T.~Aila,
  ``Analyzing and improving the image quality of stylegan,'' in
  \emph{Proceedings of the IEEE Conference on Computer Vision and Pattern
  Recognition}, 2020, pp. 8110--8119.

\bibitem{karras2021alias}
T.~Karras, M.~Aittala, S.~Laine, E.~H{\"a}rk{\"o}nen, J.~Hellsten, J.~Lehtinen,
  and T.~Aila, ``Alias-free generative adversarial networks,'' \emph{Advances
  in Neural Information Processing Systems}, vol.~34, 2021.

\bibitem{nirkin2019fsgan}
Y.~Nirkin, Y.~Keller, and T.~Hassner, ``Fsgan: Subject agnostic face swapping
  and reenactment,'' in \emph{Proceedings of the IEEE International Conference
  on Computer Vision}, 2019, pp. 7184--7193.

\bibitem{nitzan2020face}
Y.~Nitzan, A.~Bermano, Y.~Li, and D.~Cohen-Or, ``Face identity disentanglement
  via latent space mapping,'' \emph{ACM Transactions on Graphics}, vol.~39,
  no.~6, pp. 1--14, 2020.

\bibitem{chen2020simswap}
R.~Chen, X.~Chen, B.~Ni, and Y.~Ge, ``Simswap: An efficient framework for high
  fidelity face swapping,'' in \emph{Proceedings of the ACM International
  Conference on Multimedia}, 2020, pp. 2003--2011.

\bibitem{zhu2021one}
Y.~Zhu, Q.~Li, J.~Wang, C.-Z. Xu, and Z.~Sun, ``One shot face swapping on
  megapixels,'' in \emph{Proceedings of the IEEE Conference on Computer Vision
  and Pattern Recognition}, 2021, pp. 4834--4844.

\bibitem{DeepFake-url}
DeepFakes,
  \url{https://github.com/ondyari/FaceForensics/tree/master/dataset/DeepFakes},
  accessed: 2023-05-05.

\bibitem{petrov2020deepfacelab}
K.~Liu, I.~Perov, D.~Gao, N.~Chervoniy, W.~Zhou, and W.~Zhang, ``Deepfacelab:
  Integrated, flexible and extensible face-swapping framework,'' \emph{Pattern
  Recognition}, vol. 141, p. 109628, 2023.

\bibitem{blanz2004exchanging}
V.~Blanz, K.~Scherbaum, T.~Vetter, and H.-P. Seidel, ``Exchanging faces in
  images,'' in \emph{Computer Graphics Forum}, vol.~23, no.~3, 2004, pp.
  669--676.

\bibitem{bitouk2008face}
D.~Bitouk, N.~Kumar, S.~Dhillon, P.~Belhumeur, and S.~K. Nayar, ``Face
  swapping: automatically replacing faces in photographs,'' in \emph{ACM
  SIGGRAPH}, 2008, pp. 1--8.

\bibitem{lin2012face}
Y.~Lin, S.~Wang, Q.~Lin, and F.~Tang, ``Face swapping under large pose
  variations: A 3d model based approach,'' in \emph{Proceedings of the IEEE
  International Conference on Multimedia and Expo}, 2012, pp. 333--338.

\bibitem{lin2014pose}
Y.~Lin and S.~Wang, ``Pose-free face swapping based on a deformable 3d shape
  morphable model,'' \emph{IEICE Transactions on Information and Systems},
  vol.~97, no.~2, pp. 305--314, 2014.

\bibitem{cheng20093d}
Y.-T. Cheng, V.~Tzeng, Y.-Y. Chuang, and M.~Ouhyoung, ``3d morphable model
  based face replacement in video,'' in \emph{ACM SIGGRAPH}, 2009.

\bibitem{dale2011video}
K.~Dale, K.~Sunkavalli, M.~K. Johnson, D.~Vlasic, W.~Matusik, and H.~Pfister,
  ``Video face replacement,'' in \emph{Proceedings of the SIGGRAPH Asia
  Conference}, 2011, pp. 1--10.

\bibitem{nirkin2018face}
Y.~Nirkin, I.~Masi, A.~T. Tuan, T.~Hassner, and G.~Medioni, ``On face
  segmentation, face swapping, and face perception,'' in \emph{Proceedings of
  the IEEE International Conference on Automatic Face and Gesture Recognition},
  2018, pp. 98--105.

\bibitem{bao2017cvae}
J.~Bao, D.~Chen, F.~Wen, H.~Li, and G.~Hua, ``Cvae-gan: fine-grained image
  generation through asymmetric training,'' in \emph{Proceedings of the IEEE
  International Conference on Computer VSision}, 2017, pp. 2745--2754.

\bibitem{korshunova2017fast}
I.~Korshunova, W.~Shi, J.~Dambre, and L.~Theis, ``Fast face-swap using
  convolutional neural networks,'' in \emph{Proceedings of the IEEE
  International Conference on Computer Vision}, 2017, pp. 3677--3685.

\bibitem{natsume2018rsgan}
R.~Natsume, T.~Yatagawa, and S.~Morishima, ``Rsgan: Face swapping and editing
  using face and hair representation in latent spaces,'' in \emph{ACM
  SIGGRAPH}, 2018, p.~69.

\bibitem{naruniec2020high}
J.~Naruniec, L.~Helminger, C.~Schroers, and R.~Weber, ``High-resolution neural
  face swapping for visual effects,'' in \emph{Computer Graphics Forum},
  vol.~39, no.~4, 2020, pp. 173--184.

\bibitem{bao2018towards}
J.~Bao, D.~Chen, F.~Wen, H.~Li, and G.~Hua, ``Towards open-set identity
  preserving face synthesis,'' in \emph{Proceedings of the IEEE Conference on
  Computer Vision and Pattern Recognition}, 2018, pp. 6713--6722.

\bibitem{li2020advancing}
L.~Li, J.~Bao, H.~Yang, D.~Chen, and F.~Wen, ``Advancing high fidelity identity
  swapping for forgery detection,'' in \emph{Proceedings of the IEEE Conference
  on Computer Vision and Pattern Recognition}, 2020, pp. 5074--5083.

\bibitem{xu2022high}
Y.~Xu, B.~Deng, J.~Wang, Y.~Jing, J.~Pan, and S.~He, ``High-resolution face
  swapping via latent semantics disentanglement,'' in \emph{Proceedings of the
  IEEE Conference on Computer Vision and Pattern Recognition}, 2022, pp.
  7642--7651.

\bibitem{xu2022region}
C.~Xu, J.~Zhang, M.~Hua, Q.~He, Z.~Yi, and Y.~Liu, ``Region-aware face
  swapping,'' in \emph{Proceedings of the IEEE Conference on Computer Vision
  and Pattern Recognition}, 2022, pp. 7632--7641.

\bibitem{pighin2006synthesizing}
F.~Pighin, J.~Hecker, D.~Lischinski, R.~Szeliski, and D.~H. Salesin,
  ``Synthesizing realistic facial expressions from photographs,'' in \emph{ACM
  SIGGRAPH}, 2006, p.~19.

\bibitem{thies2015real}
J.~Thies, M.~Zollh{\"o}fer, M.~Nie{\ss}ner, L.~Valgaerts, M.~Stamminger, and
  C.~Theobalt, ``Real-time expression transfer for facial reenactment,''
  \emph{ACM Transactions on Graphics}, vol.~34, no.~6, pp. 183--1, 2015.

\bibitem{thies2016face2face}
J.~Thies, M.~Zollhofer, M.~Stamminger, C.~Theobalt, and M.~Nie{\ss}ner,
  ``Face2face: Real-time face capture and reenactment of rgb videos,'' in
  \emph{Proceedings of the IEEE Conference on Computer Vision and Pattern
  Recognition}, 2016, pp. 2387--2395.

\bibitem{wu2018reenactgan}
W.~Wu, Y.~Zhang, C.~Li, C.~Qian, and C.~Change~Loy, ``Reenactgan: Learning to
  reenact faces via boundary transfer,'' in \emph{Proceedings of the European
  Conference on Computer Vision}, 2018, pp. 603--619.

\bibitem{zakharov2019few}
E.~Zakharov, A.~Shysheya, E.~Burkov, and V.~Lempitsky, ``Few-shot adversarial
  learning of realistic neural talking head models,'' in \emph{Proceedings of
  the IEEE International Conference on Computer Vision}, 2019, pp. 9459--9468.

\bibitem{zhang2019one}
Y.~Zhang, S.~Zhang, Y.~He, C.~Li, C.~C. Loy, and Z.~Liu, ``One-shot face
  reenactment,'' in \emph{Proceedings of the British Machine Vision
  Conference}, 2019.

\bibitem{ha2019marionette}
S.~Ha, M.~Kersner, B.~Kim, S.~Seo, and D.~Kim, ``Marionette: Few-shot face
  reenactment preserving identity of unseen targets,'' in \emph{Proceedings of
  the AAAI Conference on Artificial Intelligence}, vol.~34, no.~07, 2020, pp.
  10\,893--10\,900.

\bibitem{burkov2020neural}
E.~Burkov, I.~Pasechnik, A.~Grigorev, and V.~Lempitsky, ``Neural head
  reenactment with latent pose descriptors,'' in \emph{Proceedings of the IEEE
  Conference on Computer Vision and Pattern Recognition}, 2020, pp.
  13\,786--13\,795.

\bibitem{gatys2016image}
L.~A. Gatys, A.~S. Ecker, and M.~Bethge, ``Image style transfer using
  convolutional neural networks,'' in \emph{Proceedings of the IEEE Conference
  on Computer Vision and Pattern Recognition}, 2016, pp. 2414--2423.

\bibitem{huang2018multimodal}
X.~Huang, M.-Y. Liu, S.~Belongie, and J.~Kautz, ``Multimodal unsupervised
  image-to-image translation,'' in \emph{Proceedings of the European Conference
  on Computer Vision}, 2018, pp. 172--189.

\bibitem{he2016deep}
K.~He, X.~Zhang, S.~Ren, and J.~Sun, ``Deep residual learning for image
  recognition,'' in \emph{Proceedings of the IEEE Conference on Computer Vision
  and Pattern Recognition}, 2016, pp. 770--778.

\bibitem{he2017mask}
K.~He, G.~Gkioxari, P.~Doll{\'a}r, and R.~Girshick, ``Mask r-cnn,'' in
  \emph{Proceedings of the IEEE International Conference on Computer Vision},
  2017, pp. 2961--2969.

\bibitem{schroff2015facenet}
F.~Schroff, D.~Kalenichenko, and J.~Philbin, ``Facenet: A unified embedding for
  face recognition and clustering,'' in \emph{Proceedings of the IEEE
  Conference on Computer Vision and Pattern Recognition}, 2015, pp. 815--823.

\bibitem{deng2019arcface}
J.~Deng, J.~Guo, N.~Xue, and S.~Zafeiriou, ``Arcface: Additive angular margin
  loss for deep face recognition,'' in \emph{Proceedings of the IEEE Conference
  on Computer Vision and Pattern Recognition}, 2019, pp. 4690--4699.

\bibitem{ronneberger2015u}
O.~Ronneberger, P.~Fischer, and T.~Brox, ``U-net: Convolutional networks for
  biomedical image segmentation,'' in \emph{International Conference on Medical
  Image Ccomputing and Computer-Assisted Intervention}, 2015, pp. 234--241.

\bibitem{yu2018bisenet}
C.~Yu, J.~Wang, C.~Peng, C.~Gao, G.~Yu, and N.~Sang, ``Bisenet: Bilateral
  segmentation network for real-time semantic segmentation,'' in
  \emph{Proceedings of the European Conference on Computer Vision}, 2018, pp.
  325--341.

\bibitem{song2018geometry}
L.~Song, Z.~Lu, R.~He, Z.~Sun, and T.~Tan, ``Geometry guided adversarial facial
  expression synthesis,'' in \emph{Proceedings of the ACM International
  Conference on Multimedia}, 2018, pp. 627--635.

\bibitem{bulat2017far}
A.~Bulat and G.~Tzimiropoulos, ``How far are we from solving the 2d \& 3d face
  alignment problem?'' in \emph{Proceedings of the IEEE International
  Conference on Computer Vision}, 2017, pp. 1021--1030.

\bibitem{park2019semantic}
T.~Park, M.-Y. Liu, T.-C. Wang, and J.-Y. Zhu, ``Semantic image synthesis with
  spatially-adaptive normalization,'' in \emph{Proceedings of the IEEE
  Conference on Computer Vision and Pattern Recognition}, 2019, pp. 2337--2346.

\bibitem{liu2015faceattributes}
Z.~Liu, P.~Luo, X.~Wang, and X.~Tang, ``Deep learning face attributes in the
  wild,'' in \emph{Proceedings of the IEEE International Conference on Computer
  Vision}, 2015, pp. 3730--3738.

\bibitem{karras2018progressive}
T.~Karras, T.~Aila, S.~Laine, and J.~Lehtinen, ``Progressive growing of gans
  for improved quality, stability, and variation,'' in \emph{International
  Conference on Learning Representations}, 2018.

\bibitem{FaceSwap-url}
FaceSwap,
  \url{https://github.com/ondyari/FaceForensics/tree/master/dataset/FaceSwapKowalski},
  accessed: 2023-05-05.

\bibitem{thies2019deferred}
J.~Thies, M.~Zollh{\"o}fer, and M.~Nie{\ss}ner, ``Deferred neural rendering:
  Image synthesis using neural textures,'' \emph{ACM Transactions on Graphics},
  vol.~38, no.~4, pp. 1--12, 2019.

\bibitem{paszke2019pytorch}
A.~Paszke, S.~Gross, F.~Massa, A.~Lerer, J.~Bradbury, G.~Chanan, T.~Killeen,
  Z.~Lin, N.~Gimelshein, L.~Antiga \emph{et~al.}, ``Pytorch: An imperative
  style, high-performance deep learning library,'' in \emph{Advances in Neural
  Information Processing Systems}, 2019, pp. 8026--8037.

\bibitem{choi2020stargan}
Y.~Choi, Y.~Uh, J.~Yoo, and J.-W. Ha, ``Stargan v2: Diverse image synthesis for
  multiple domains,'' in \emph{Proceedings of the IEEE Conference on Computer
  Vision and Pattern Recognition}, 2020, pp. 8188--8197.

\bibitem{heusel2017gans}
M.~Heusel, H.~Ramsauer, T.~Unterthiner, B.~Nessler, and S.~Hochreiter, ``Gans
  trained by a two time-scale update rule converge to a local nash
  equilibrium,'' \emph{Advances in Neural Information Processing Systems},
  vol.~30, 2017.

\bibitem{wang2018cosface}
H.~Wang, Y.~Wang, Z.~Zhou, X.~Ji, D.~Gong, J.~Zhou, Z.~Li, and W.~Liu,
  ``Cosface: Large margin cosine loss for deep face recognition,'' in
  \emph{Proceedings of the IEEE Conference on Computer Vision and Pattern
  Recognition}, 2018, pp. 5265--5274.

\bibitem{doosti2020hope}
B.~Doosti, S.~Naha, M.~Mirbagheri, and D.~J. Crandall, ``Hope-net: A
  graph-based model for hand-object pose estimation,'' in \emph{Proceedings of
  the IEEE Conference on Computer Vision and Pattern Recognition}, 2020, pp.
  6608--6617.

\bibitem{deng2019accurate}
Y.~Deng, J.~Yang, S.~Xu, D.~Chen, Y.~Jia, and X.~Tong, ``Accurate 3d face
  reconstruction with weakly-supervised learning: From single image to image
  set,'' in \emph{Proceedings of the IEEE Conference on Computer Vision and
  Pattern Recognition Workshops}, 2019, pp. 285--295.

\bibitem{szegedy2016rethinking}
C.~Szegedy, V.~Vanhoucke, S.~Ioffe, J.~Shlens, and Z.~Wojna, ``Rethinking the
  inception architecture for computer vision,'' in \emph{Proceedings of the
  IEEE Conference on Computer Vision and Pattern Recognition}, 2016, pp.
  2818--2826.

\bibitem{MegaFace-url}
MegaFace,
  \url{https://github.com/zyainfal/One-Shot-Face-Swapping-on-Megapixels},
  accessed: 2023-05-05.

\bibitem{FSGAN-url}
FSGAN, \url{https://github.com/YuvalNirkin/fsgan}, accessed: 2023-05-05.

\bibitem{FSLSD-url}
FSLSD, \url{https://github.com/cnnlstm/FSLSD_HiRes}, accessed: 2023-05-05.

\bibitem{RAFSwap-url}
RAFSwap, \url{https://github.com/xc-csc101/RAFSwap}, accessed: 2023-05-05.

\bibitem{shen2020interfacegan}
Y.~Shen, C.~Yang, X.~Tang, and B.~Zhou, ``Interfacegan: Interpreting the
  disentangled face representation learned by gans,'' \emph{IEEE Transactions
  on Pattern Analysis and Machine Intelligence}, vol.~44, no.~4, pp.
  2004--2018, 2020.

\bibitem{shen2021closed}
Y.~Shen and B.~Zhou, ``Closed-form factorization of latent semantics in gans,''
  in \emph{Proceedings of the IEEE Conference on Computer Vision and Pattern
  Recognition}, 2021, pp. 1532--1540.

\bibitem{zhang2016joint}
K.~Zhang, Z.~Zhang, Z.~Li, and Y.~Qiao, ``Joint face detection and alignment
  using multitask cascaded convolutional networks,'' \emph{IEEE Signal
  Processing Letters}, vol.~23, no.~10, pp. 1499--1503, 2016.

\end{thebibliography}

\vspace{-12mm}
\begin{IEEEbiography}[{\includegraphics[width=1in,height=1.25in,clip,keepaspectratio]{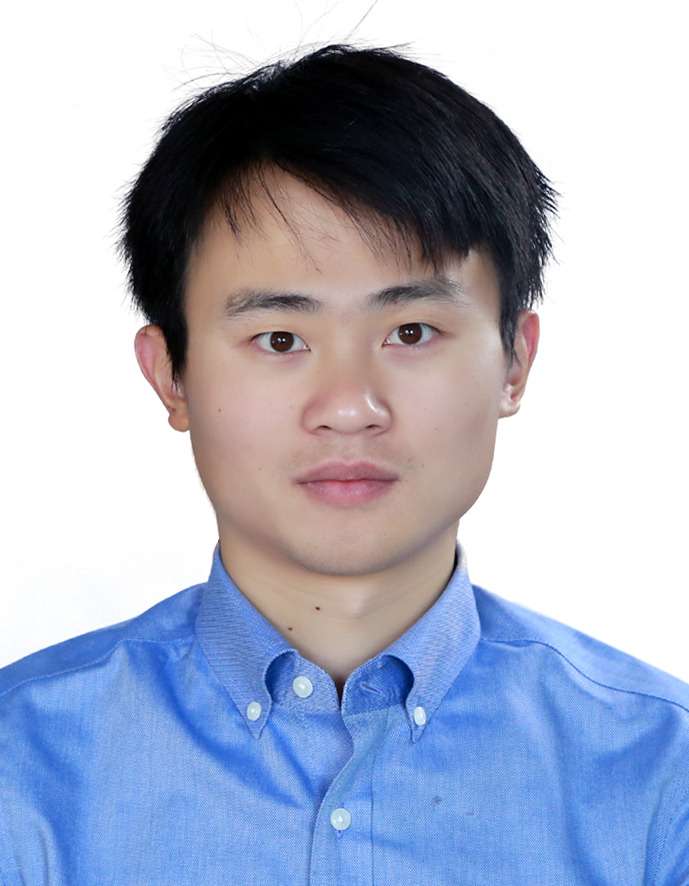}}]{Qi Li}
  received the B.E. degree from China University of Petroleum in 2011, the Ph.D. degree from the Institute of Automation, Chinese Academy of Sciences (CASIA) in 2016.
  He is an Associate Professor with the New Laboratory of Pattern Recognition (NLPR), State Key Laboratory of Multimodal Artificial Intelligence Systems (MAIS), CASIA. 
  His research interests include face recognition, computer vision, and machine learning.
\end{IEEEbiography}

\vspace{-12mm}
\begin{IEEEbiography}[{\includegraphics[width=1in,height=1.25in,clip,keepaspectratio]{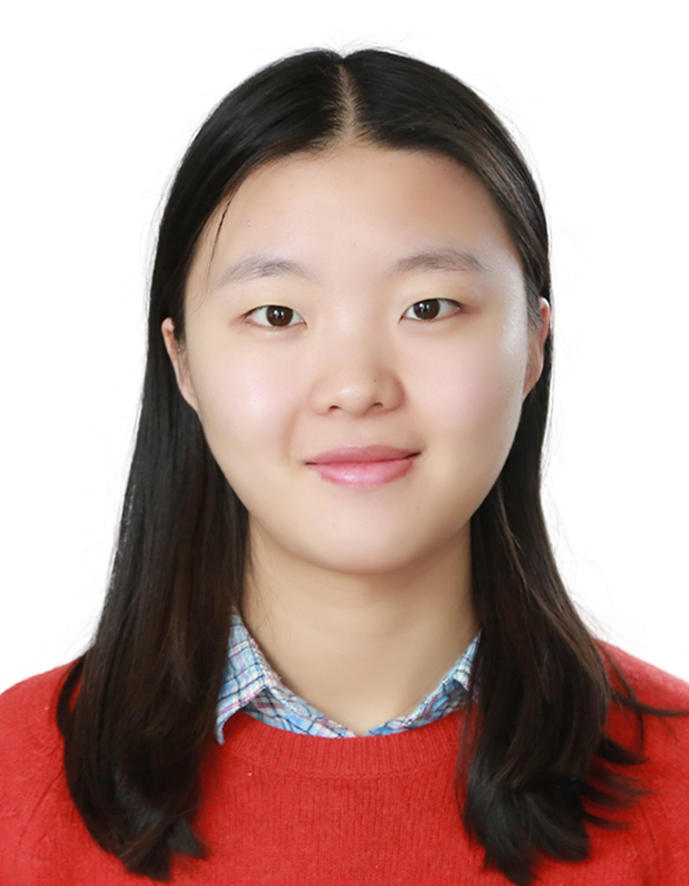}}]{Weining Wang}
  received her B.E. degree from North China Electric Power University in 2015 and the Ph.D. degree from University of Chinese Academy of Sciences (UCAS) in 2020. She is now an Associate Professor at the Laboratory of Cognition and Decision Intelligence for Complex Systems, Institute of Automation, Chinese Academy of Sciences (CASIA). Her research interests include pattern recognition and computer vision.
\end{IEEEbiography}

\vspace{-12mm}
\begin{IEEEbiography}[{\includegraphics[width=1in,height=1.25in]{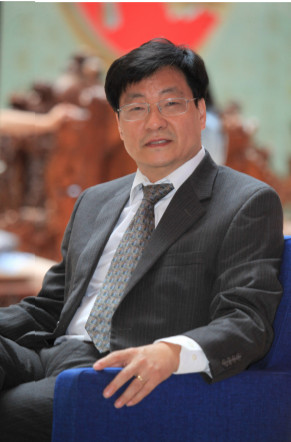}}]{Chengzhong Xu} obtained the B.E. and M.S. degrees from Nanjing University in 1986, and 1989, respectively, and the Ph.D. degree from the University of Hong Kong in 1993, all in computer science and engineering.
He is now a Chair Professor of Computer Science at the State Key Laboratory of Internet of Things for Smart City and the Dean of Faculty of Science and Technology, University of Macau, Macao SAR, China.  
\end{IEEEbiography}

\vspace{-12mm}
\begin{IEEEbiography}[{\includegraphics[width=1in,height=1.25in,clip,keepaspectratio]{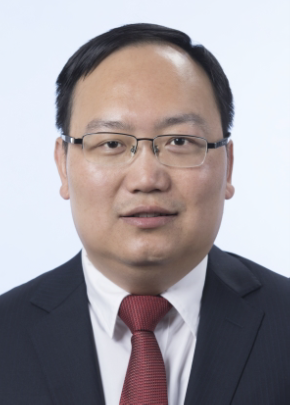}}]{Zhenan Sun}
  received the Ph.D. degree from the Institute of Automation, Chinese Academy of Sciences (CASIA) in 2006.
  He is a professor at New Laboratory of Pattern Recognition (NLPR), State Key Laboratory of Multimodal Artificial Intelligence Systems (MAIS), CASIA.
  His current research interests include biometrics, pattern recognition, and computer vision. He is a fellow of the IAPR, and an Associate Editor of the IEEE Transactions on Biometrics, Behavior, and Identity Science.
\end{IEEEbiography}

\vspace{-12mm}
\begin{IEEEbiography}[{\includegraphics[width=1in,height=1.25in,clip,keepaspectratio]{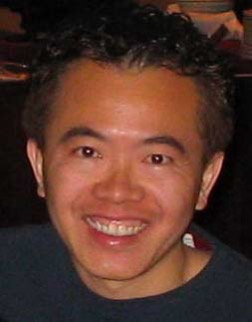}}]{Ming-Hsuan Yang}
  is affiliated with University of California at Merced and Yonsei University. 
  He served as program co-chair of ICCV 2019 and co-editor-in-chief of CVIU.
  He received Longuet-Higgins Prize in 2023, Best Paper Honorable Mention at CVPR 2018, NSF CAREER Award in 2012, and Google Faculty Award in 2009. 
  He is a fellow of the IEEE and ACM.
\end{IEEEbiography}

\end{document}